# Title

Primate vision reveals a missing principle for robust dynamic AI

## Authors

Matteo Dunnhofer[1,2], Christian Micheloni[2], Kohitij Kar[1]

## Affiliation

1. York University, Department of Biology and Centre for Vision Research, Toronto, Canada
2. Department of Mathematics, Computer Science, and Physics, University of Udine, Udine, Italy

* Correspondence should be addressed to Matteo Dunnhofer, Kohitij Kar
E-mail: matteo24@yorku.ca , k0h1t1j@yorku.ca

## Conflict of interests

The authors declare no competing financial interests.

## Acknowledgments

KK has been supported by funds from the Canada Research Chair Program (CRC-2021-00326), the Canada First Research Excellence Funds (VISTA Program), and the National Sciences and Engineering Research Council of Canada (NSERC, RGPIN-2024-06223). MD received funding from the European Union's Horizon Europe research and innovation programme under the Marie Skłodowska- Curie grant agreement n. 101151834 (PRINNEVOT).

# Abstract

How does an intelligent visual system combine what objects look like with how they move while remaining robust as appearance changes? We addressed this question by comparing human perception and neural activity in macaque inferior temporal cortex with representations from image- and video-based neural networks spanning recognition, segmentation, optic-flow processing and predictive world modeling. Temporal integration improved object representations, but most video recognition models generalized poorly when appearance was disrupted while motion structure was preserved. Humans and macaque IT remained robust. Notably, predictive world models combined strong cross-appearance generalization with the closest correspondence to IT, outperforming other video-modeling approaches in neural fidelity. Yet no model reproduced the cortical transformation from early appearance-dominated responses toward later appearance-invariant motion coding. These results identify progressive integration of motion into object representations as a principle of robust dynamic vision and implicate predictive learning as a promising route toward realizing this computation in artificial systems.

# Introduction

Intelligent vision must operate in a world that is continuously changing. Objects move, viewpoints vary and visual evidence unfolds over time, yet observers maintain stable representations of what objects are while simultaneously extracting how they are moving. Artificial neural networks (ANNs) optimized for static image recognition have provided powerful computational models of primate object recognition and have predicted aspects of both behavior and neural activity along the ventral visual stream [1–4]. But natural vision is fundamentally dynamic[5–9] , raising an important question for both artificial intelligence and neuroscience: what computations allow a visual system to transform changing sensory input into robust representations of objects and their dynamics? Video-based neural networks offer an intuitive answer. By processing multiple video frames jointly, these models can integrate information across space and time rather than treating each moment independently. Such architectures have achieved impressive performance in video recognition, object tracking, and related dynamic visual tasks[10–18]. Temporal integration should be particularly valuable when the relevant information is distributed across time: object identity may often be inferred from individual frames, whereas properties such as motion direction and speed necessarily depend on relationships between successive observations. This has motivated the idea that extending successful image-based recognition systems with spatiotemporal processing may provide a computational account of dynamic visual intelligence.

High performance on video benchmarks, however, does not necessarily imply robust understanding of object dynamics. Natural videos contain strong correlations between what an object looks like and how it moves. A model can therefore improve at video tasks by integrating appearance information over time without developing a representation of motion that generalizes when appearance changes[19]. This distinction is consequential. A robust dynamic representation should preserve useful information about object appearance while also extracting motion structure in a form that can be read out across variations in that appearance. Standard video benchmarks rarely isolate these components, making it difficult to determine whether temporal integration produces genuinely dynamic representations or simply provides additional opportunities to exploit appearance-dependent cues.

Existing evidence leaves this question unresolved. Video ANNs outperform image-based models on many dynamic computer-vision benchmarks[10–18], and representations learned from video have shown improved correspondence with cortical activity during naturalistic viewing[20]. Yet naturalistic performance alone cannot identify the computations responsible for this improvement. Moreover, neural correspondence measured using fMRI may provide limited access to the rapid temporal transformations underlying dynamic visual processing because of the comparatively slow timescale of the hemodynamic response[21]. Thus, despite substantial advances in video modeling, it remains unclear whether current architectures reproduce the representational transformations that make biological dynamic vision robust.

Biological vision provides a useful computational reference for resolving this ambiguity. The primate ventral visual stream is traditionally associated with constructing representations that support object recognition[22–24], but recent work demonstrates that the inferior temporal (IT) cortex also contains information about object motion[7]. This suggests that high-level object representations are not purely static descriptions of visual form. Instead, representations within the ventral stream may evolve as visual evidence accumulates, integrating information about both object identity and object dynamics. Critically, this possibility predicts more than the presence of motion information in IT. If the brain constructs robust dynamic representations, motion information should increasingly become accessible across changes in object appearance rather than remaining bound to the visual features from which it was initially derived.

This biological perspective also raises a question about how artificial systems might acquire such representations. Conventional video-recognition networks introduce temporal operations into architectures optimized to classify video content [10–12,15,25]. Other approaches impose more explicit structure on motion processing: optic-flow-based systems directly represent displacement between successive frames[18,26]. More recently, predictive video world models learn representations by using the temporal structure of latent visual experience itself as a supervisory signal[17]. Rather than relying exclusively on semantic recognition objectives, these models are trained to capture information useful for predicting missing or future states of a dynamic scene. Such predictive learning could, in principle, encourage representations that capture persistent spatiotemporal structure across changes in instantaneous appearance. Whether this produces representations more similar to the biological solution, however, remains unknown.

Here we test these possibilities by placing human perception, macaque IT neural activity and diverse computational models within the same controlled framework. We first compare image-based and video-recognition ANNs on dynamic object tasks spanning object identity, motion direction and motion speed. We then introduce an appearance-free manipulation that preserves object motion while disrupting recognizable visual form[19]. This manipulation provides a stringent test of whether motion information extracted from naturalistic videos generalizes beyond the appearance statistics on which it was learned. We extend this benchmark across video-recognition networks, object-segmentation systems, models with explicit optic-flow processing[18,26] and predictive video world models[17].

We next ask whether the computational strategies that support robust behavior also account for neural representations in the primate ventral stream. Using large-scale recordings from macaque IT while animals viewed the same dynamic stimuli, we test whether motion representations generalize across appearance disruption and whether different classes of video ANNs capture the temporal evolution of IT population activity. Finally, we move beyond population-level correspondence to examine the dynamics of individual neural and model units, asking whether biological and artificial representations undergo similar transformations in their relative sensitivity to object appearance and motion.

Our results reveal an important distinction between integrating information over time and constructing a robust dynamic representation. Video-recognition models benefit substantially from temporal integration, but this improvement does not generally produce motion representations that transfer across appearance changes. Human perception and macaque IT show much greater robustness. Models with explicit motion processing and predictive world modeling close part of this behavioral gap, while predictive world models provide the strongest correspondence with IT among the video-modeling approaches tested. Nevertheless, even these models fail to reproduce a defining property of the neural responses: IT representations progressively evolve from an initially appearance-dominated state toward a state in which object motion is increasingly represented across changes in appearance. Together, these findings identify progressive integration of appearance and motion into appearance-robust object representations as a computational target for dynamic AI, and point to predictive learning as a promising route toward acquiring such representations.

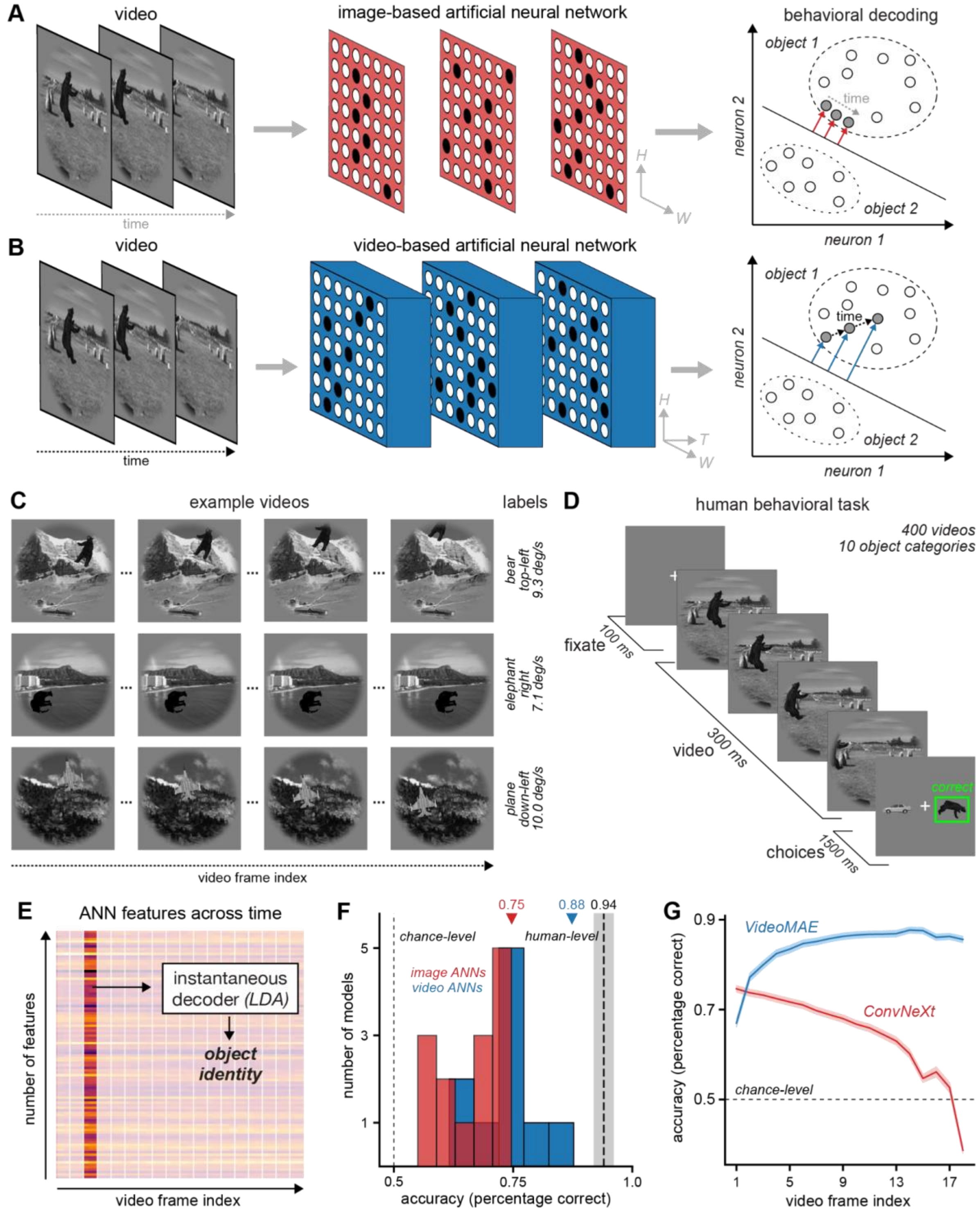

**Figure 1. Testing image and video recognition artificial neural networks (ANNs) as models of human dynamic visual object recognition. A-B.** Conceptual comparison between image-based and video-based ANNs. Dynamic stimuli in the form of videos were presented to both classes of neural network models. **A.** Image-based ANNs. Each video frame is processed independently, producing frame-wise feature representations that are object-selective but not integrated across time. The absence of temporal context prevents these networks from integrating evidence across time to disambiguate the visual input, limiting their ability of separating objects into behavioral manifolds. **B.** Video recognition ANNs. Multiple frames are processed jointly through spatiotemporal computations that integrate information across space and time. Temporal integration is expected to transform dynamic visual input into more separable behavioral manifolds, enabling improved decoding of task-relevant object variables. **C.** Example visual stimuli. Representative video frame sequences illustrating the

dynamic object stimuli used throughout the study. Videos varied in object category, motion direction, and motion speed while preserving naturalistic visual structure. **D.** Human behavioral task. Human participants viewed brief naturalistic videos (300 ms) depicting objects from 10 categories moving in different directions and at different speeds. After each video, participants reported object identity through forced-choice responses. **E-G.** Instantaneous behavioral decoding. **E.** For each ANN model, feature representations were extracted at every video frame and used to train an instantaneous linear decoder (Linear Discriminant Analysis, LDA). Decoding was performed independently at each time point to evaluate how behavioral information emerges over the course of the video. **F.** Decoding results. Decoding accuracy across the population of image-based ANNs (red) and video-based ANNs (blue) for object identity. Triangles indicate the best-performing model in each class, while the gray triangle denotes human behavioral performance. **G.** Time-resolved decoding accuracy for object identity of the best image-based ANN (ConvNeXt) compared with the best video recognition ANN (VideoMAE). Mean accuracy and standard error over videos are reported at the different video frames.

# Results

## Humans accurately discriminate moving objects

Humans can rapidly extract multiple object attributes from brief dynamic visual inputs, including identity, motion direction and speed. To establish behavioral benchmarks for these complementary dimensions of dynamic perception, participants viewed 300 ms video clips and reported object identity, motion direction or motion speed using binary-choice interfaces (**Fig. 1C-D**, **Fig. 2A-D**). The object recognition and motion direction tasks involved 400 videos (40 videos per object, 10 object categories and 8 motion directions, evenly distributed), whereas the speed discrimination task involved 320 videos (32 videos per object, 10 object categories, and 40 motion speeds).

Humans (N = 107) performed at high accuracy across all tasks (**Fig. 1F**, **Fig. 2B-E**). Object identity was discriminated with near-ceiling performance (*mean accuracy = 0.94 ± 0.02 SD*, **Fig. 1F**), motion direction was also accurately estimated (*mean accuracy = 0.91 ± 0.07 SD*, **Fig. 2B**), and speed discrimination was more challenging but remained robust (*mean accuracy = 0.82 ± 0.06 SD*, **Fig. 2E**). Thus, humans rapidly extract motion direction and speed while maintaining highly accurate object-identity judgments from the same brief dynamic inputs. These measurements establish behavioral benchmarks for asking whether artificial and neural representations make complementary object and motion information accessible during dynamic viewing.

## Temporal information improves dynamic object representations in video models

We next asked whether access to temporal context improves the representation of identity and motion information in artificial systems. We evaluated image- and video-recognition ANNs on the same 300 ms stimuli used for the human benchmarks. Image-based ANNs processed each of the 18 video frames independently, whereas video-recognition ANNs integrated information across multiple frames through spatiotemporal operations (**Fig. 1A-B**). This contrast allowed us to test whether temporal context increases the accessibility of object identity and, more critically, of dynamic variables such as motion direction and speed.

To test this hypothesis, we linearly decoded the same object behavioral variables used for humans from ANN features. For video recognition models (N = 10 models, selected among the earliest popular architectures trained on video recognition[11,12]), features indexed by each video frame were computed by integrating information over a window on the preceding frames, yielding spatiotemporally-integrated representations. For image-based models (N = 14 models, chosen based on their alignment with human behavior and ventral stream neural representation), each frame was processed independently, resulting in temporally unintegrated representations. Ideally, just as image-based ANNs optimized for object recognition have been shown to predict both human behavior and neural activity during static object recognition[2–4], a reasonable hypothesis is that video-based ANNs optimized for dynamic visual tasks (e.g., video recognition) could capture human dynamic perception and provide mechanistic models of the neural representations that support it. For each model, we trained linear decoders to predict object identity (**Fig. 1D-G**), object motion direction (**Fig. 2B-C**), and object motion speed (**Fig. 2E-F**) from the features at every frame (**Fig. 1E**). Decoder performance was evaluated using cross-validation on held-out data, taking the maximum value across video frames (**Fig. 1G, Fig. 2C, Fig. 2F**), allowing a direct comparison between model predictions and human behavioral performance.

Video recognition ANNs outperformed image-based models on the dynamic object recognition task. The accuracy of the video ANN population (*mean = 0.73 ± 0.07 SD, max = 0.88*) was significantly higher (Δ *mean accuracy = 0.07, p = 0.04, t(22) = -2.16, unpaired t-test;* **Fig. 1F**) than that of the image-based ANN population (*mean = 0.66 ± 0.07 SD, max = 0.75*). This result supports the hypothesis that the spatiotemporal processing of video ANNs improves the representation of dynamic visual stimuli. By combining information across successive frames, video ANNs can accumulate visual evidence over time and resolve ambiguities that may be difficult to overcome from individual frames alone. As a consequence, these models construct feature representations that more effectively preserve object identity throughout the dynamic stimulus. This evidence is further supported by the improvement brought by video recognition ANNs on behavioral tasks with variables explicitly depending on time integration, such as object motion direction and object motion speed. For motion direction decoding, video ANNs achieved higher accuracy (*mean = 0.65 ± 0.06 SD, max = 0.81*) compared to image-based ANNs (*mean = 0.58 ± 0.02 SD, max = 0.62;* Δ *mean accuracy = 0.07, p = 0.001, t(22) = -3.66, unpaired t-test;* **Fig. 2B**), and a similar advantage was observed for speed decoding (video ANNs: *mean = 0.67 ± 0.04 SD, max = 0.76;* image-based ANNs: *mean = 0.63 ± 0.02 SD, max = 0.66*; Δ *mean accuracy = 0.04, p = 0.02, t(22) = -2.49, unpaired t-test;* **Fig. 2F**), making them closer to human behavioral performance.

Because image- and video-based networks also differ in architecture, training objectives and training data, this cross-model comparison alone cannot isolate the contribution of temporal information. We therefore performed a within-architecture temporal-input ablation on the video-recognition models. For each network, we compared its standard spatiotemporal representation with a matched condition in which each frame was processed without temporally varying input by repeating the current frame across the model's temporal input window. Removing temporal variation significantly reduced the linear decodability of object identity, motion direction and motion speed (identity: Δ (spatiotemporal - spatial) *mean accuracy = 0.11, p < 0.001, t(9) = 4.85, paired t-test;* motion direction: Δ (spatiotemporal - spatial) *mean accuracy = 0.10, p = 0.002, t(9) = 4.25, paired t-test;* motion speed: Δ (spatiotemporal - spatial) *mean accuracy = 0.06, p = 0.003, t(9) = 4.02, paired t-test;* **Fig. S1**). Thus, access to information distributed across successive frames contributes directly to the representational advantage of video-recognition models, although this manipulation does not imply that temporal processing is the only difference responsible for their advantage over independently trained image models.

Together, these analyses show that temporally varying input increases the accessibility of identity and motion-related information in video-model representations. However, successful decoding of a motion variable does not necessarily imply a robust representation of motion itself. Because motion and appearance covary in natural videos, we next asked whether this information generalizes when motion is preserved but recognizable object appearance is removed.

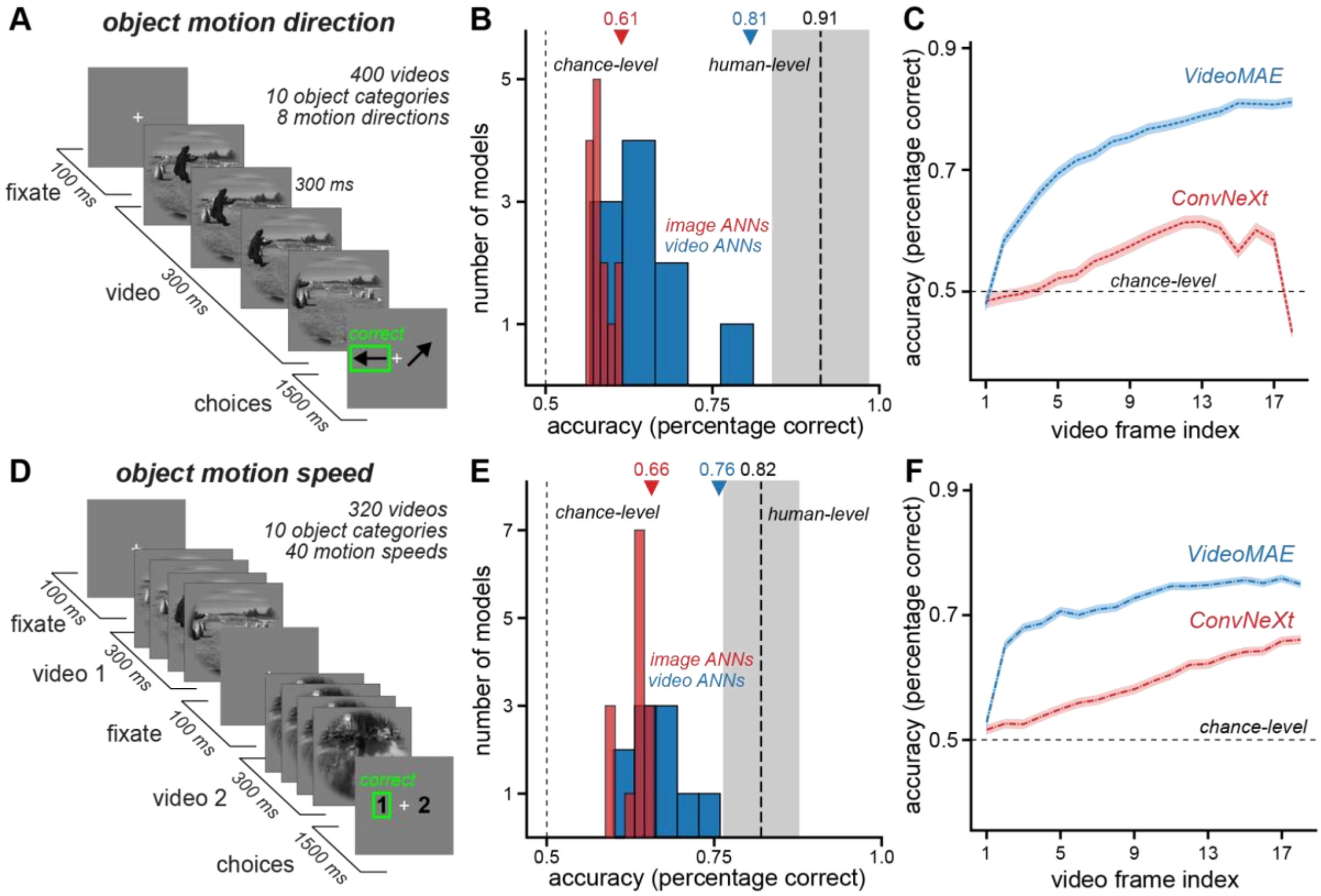

**Figure 2. Behavioral comparison of image and video recognition ANNs on object motion tasks.** **A.** Motion-direction behavioral task. Human observers viewed short naturalistic videos and reported the direction of object motion. Each trial consisted of a 100 ms fixation period, a 300 ms video presentation, and a response phase of 1500 ms. **B.** Distribution of motion-direction decoding performance across models. Histograms show the maximum decoding accuracy achieved by each image-based ANN (red) and video-based ANN (blue). Triangles indicate the highest-performing model in each group. **C.** Time-resolved decoding of motion direction from ANN representations. For each video frame, linear decoders were trained on model features to predict object motion direction. Curves show decoding accuracy across video frames for the top-performing object recognition image-based model (ConvNeXt, red) and the top-performing video recognition model (VideoMAE, blue). Mean accuracy and standard error over videos are reported at the different video frame indices. **D.** Motion speed behavioral task. Human observers viewed pairs of videos depicting different object motions and judged which object moved faster. Each trial consisted of two sequential video presentations followed by a speed-comparison response. **E.** Distribution of speed-decoding performance across models. Histograms show peak decoding accuracy across video frames for all models. **F.** Time-resolved decoding of object speed from ANN representations. Using the same decoding procedure as in (**C**), object speed was predicted from model features at each video frame.

## Video recognition models encode motion but show limited generalization beyond appearance

The preceding analyses showed that motion direction can be decoded from both image- and video-model representations under naturalistic viewing. Yet natural videos contain appearance cues correlated with motion, leaving open whether this decodability reflects motion information that generalizes across changes in appearance or information tied to the visual features present in the training condition [7,19]. This distinction is critical because standard decoding under naturalistic conditions cannot, by itself, distinguish an appearance-bound correlate of motion from an appearance-robust dynamic representation.

We first asked how robust human motion perception is to the removal of such structured appearance information. To test this, we designed *appearance-free* stimuli (N = 100 videos, 10 videos per object, 10 object categories, 8 motion directions) in which each video frame consisted of random pixel patterns, while a subset of pixels was coherently displaced across frames to match the motion of an underlying object. These stimuli were generated by extracting pixel-wise motion from naturalistic videos using optic flow and applying it to noise images, thereby preserving motion signals while eliminating object appearance[19] (**Fig. 3A-D**). Participants viewed these videos and reported the perceived motion direction using a two-choice interface. As a control, we also measured performance on the corresponding appearance-based stimuli, where the original visual content was preserved. Despite the absence of appearance cues, human performance remained high (*mean accuracy = 0.90* ± *0.06 SD*; **Fig. 3D**), approximately equalizing performance in the appearance-based condition (*mean accuracy = 0.92* ± *0.04 SD*; **Fig. 3B**). Thus, human motion judgments remain highly robust when recognizable appearance is removed, establishing a stringent cross-appearance benchmark for artificial representations.

We next applied the same test to image- and video-recognition ANNs. Motion-direction decoders were trained only on features from naturalistic videos and, without refitting, tested on both held-out naturalistic representations and their paired appearance-free representations [7]. Decoding was high in the naturalistic condition for both video-recognition ANNs (*mean accuracy = 0.78* ± *0.09 SD*; **Fig. 3B**) and image-based ANNs (*mean accuracy = 0.71* ± *0.05 SD*; **Fig. 3B**). Cross-appearance transfer, however, was markedly weaker. Image-based ANNs fell to chance-level performance (*mean accuracy = 0.47* ± *0.02 SD*; **Fig. 3D**), showing that motion-direction information available from their naturalistic representations did not transfer when appearance was removed. Some video-recognition ANNs retained above-chance information, but the model class remained far below the human benchmark (*mean accuracy = 0.53* ± *0.05 SD*; **Fig. 3D**). Critically, therefore, high motion decodability under naturalistic viewing - even in temporally integrated models - did not imply cross-appearance generalization. Temporal integration improves access to motion-related information, but conventional video-recognition objectives do not generally produce appearance-robust motion representations.

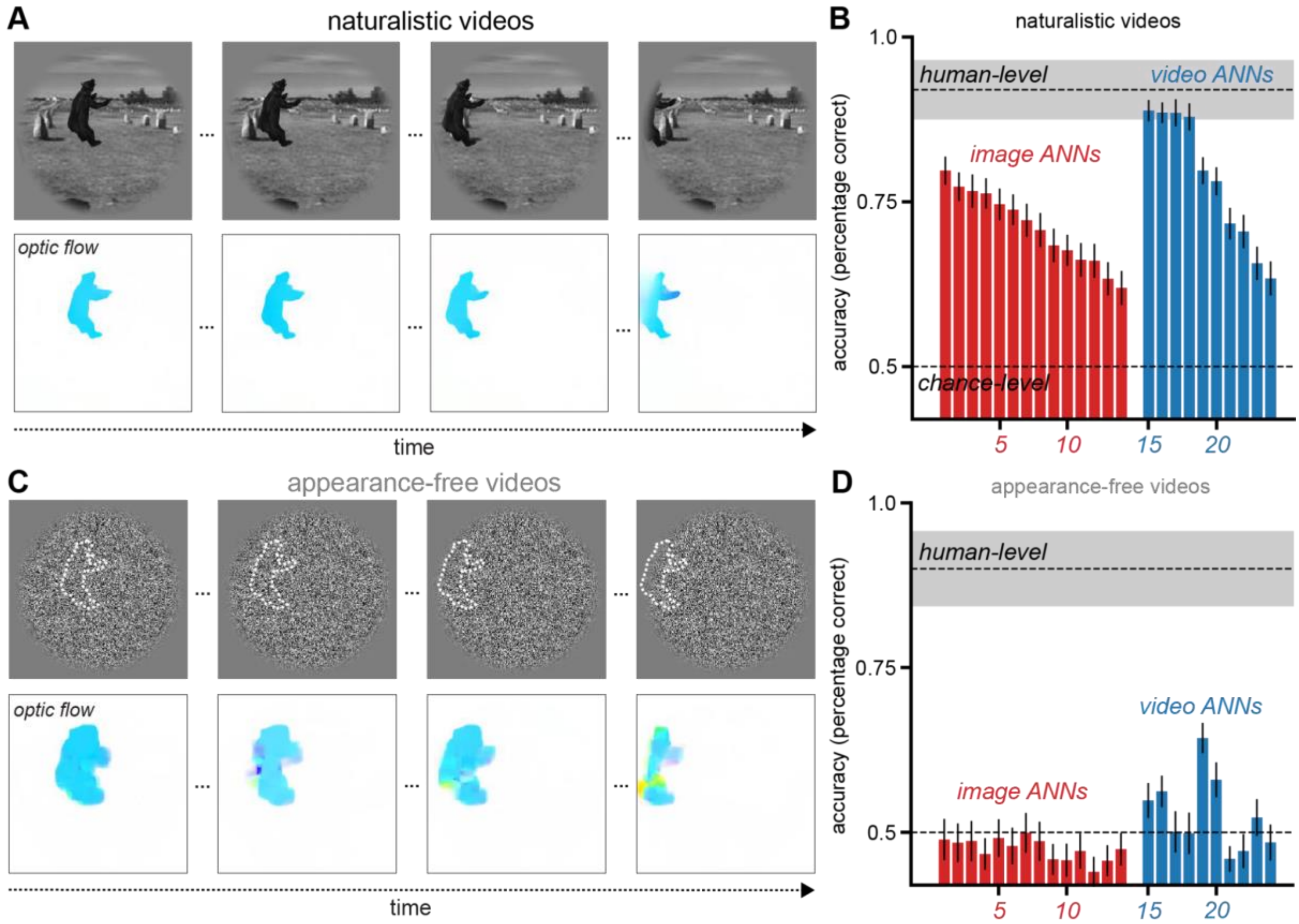

**Figure 3. Exposing robustness limitations in image and video recognition ANNs. A.** Example naturalistic video sequence (top) and the corresponding optic flow fields computed between consecutive frames (bottom). In naturalistic videos, motion information is accompanied by object appearance cues, allowing motion direction to potentially be inferred from both dynamic and static visual features. **B.** Decoding accuracy for object motion direction using representations from naturalistic videos. Linear decoders were trained and tested on representations extracted from the naturalistic stimulus condition. The horizontal line reports mean human behavioral performance with the gray area indicating standard deviation. Vertical bars indicate the maximum frame-indexed mean accuracy across videos. Error bars indicate the standard error across videos at the selected frame. Model legend - image ANNs - 1: PNASNet, 2: NASNet, 3: Inception-V3, 4: ConvNeXt, 5: DenseNet-121, 6: CORnet-S, 7: ConvRNN, 8: EfficientNet-B0, 9: AlexNet, 10: ResNet-50-SSL; 11: ViT, 12: ResNet-50, 13: ViT-SSL, 14: Hiera; Video ANNs - 15: X3D-XS, 16: I3D, 17: SlowR50-ssv2, 18: SlowR50, 19: C2D, 20: VideoMAE, 21: TimesFormer-ssv2, 22: VideoMamba-ssv2, 23: TimesFormer, 24: VideoMamba. **C.** Example appearance-free video sequence generated from the same motion trajectory shown in (**A**) and the corresponding optic flow fields. Appearance-free videos preserve motion dynamics while removing recognizable object appearance, allowing motion information to be dissociated from appearance cues. **D.** Generalization of motion-direction decoding across appearance changes. Linear decoders were trained on representations from naturalistic videos and tested on appearance-free videos. The horizontal line reports mean human behavioral performance with the gray area indicating standard deviation. Bars indicate the maximum frame-indexed mean accuracy across videos. Error bars indicate the standard error across videos at the selected frame.

## Explicit motion processing and predictive world modeling improve cross-appearance generalization

The limited transfer of conventional video-recognition models prompted us to ask whether alternative computational strategies produce more robust motion representations. We contrasted two conceptually distinct routes. Optic-flow-based video object-segmentation models (MatNet[26], FusionSeg[18]), explicitly expose inter-frame displacement signals, whereas predictive video world models (V-JEPA2 models[17]) learn spatiotemporal representations by predicting masked or future visual content. We also tested SAM2[13] as a modern appearance-based video-segmentation comparison. Optic-flow models combine motion signals computed between successive frames [27] with RGB appearance information (**Fig. 4A**), whereas world models use the temporal structure of video itself as a learning signal (**Fig. 4B**).

Using the same cross-condition decoding protocol, we trained motion-direction decoders on naturalistic representations and tested them on both naturalistic and appearance-free stimuli. Optic-flow models (MatNet, FusionSeg) and predictive world models (V-JEPA2g, V-JEPA2l) both approached the human benchmark across the two conditions (**Fig. 4C**). Optic-flow models achieved *mean accuracy = 0.91 ± 0.01 SD* on naturalistic videos and *mean accuracy = 0.79 ± 0.01 SD* on appearance-free videos; world models achieved *mean accuracy = 0.92 ± 0.02 SD* and *mean accuracy = 0.85 ± 0.04 SD*, respectively. SAM2 showed high decoding on naturalistic videos (*mean accuracy = 0.88 ± 0.03 SD*) but substantially poorer transfer to appearance-free stimuli (*mean accuracy = 0.28 ± 0.03 SD*). Thus, robust cross-appearance motion information was not shared by all temporally processing systems, but emerged much more strongly in models with explicit motion inputs or predictive world-modeling objectives.

These results show that cross-appearance motion generalization is not a generic consequence of temporal processing alone. Two different computational strategies - explicit motion representation and predictive learning from video - can both produce substantially more robust motion information than conventional video-recognition models. Behavioral robustness alone, however, cannot determine whether these strategies arrive at a representation resembling the biological solution. We therefore turned to macaque IT, where object and motion information are jointly represented during dynamic viewing [7].

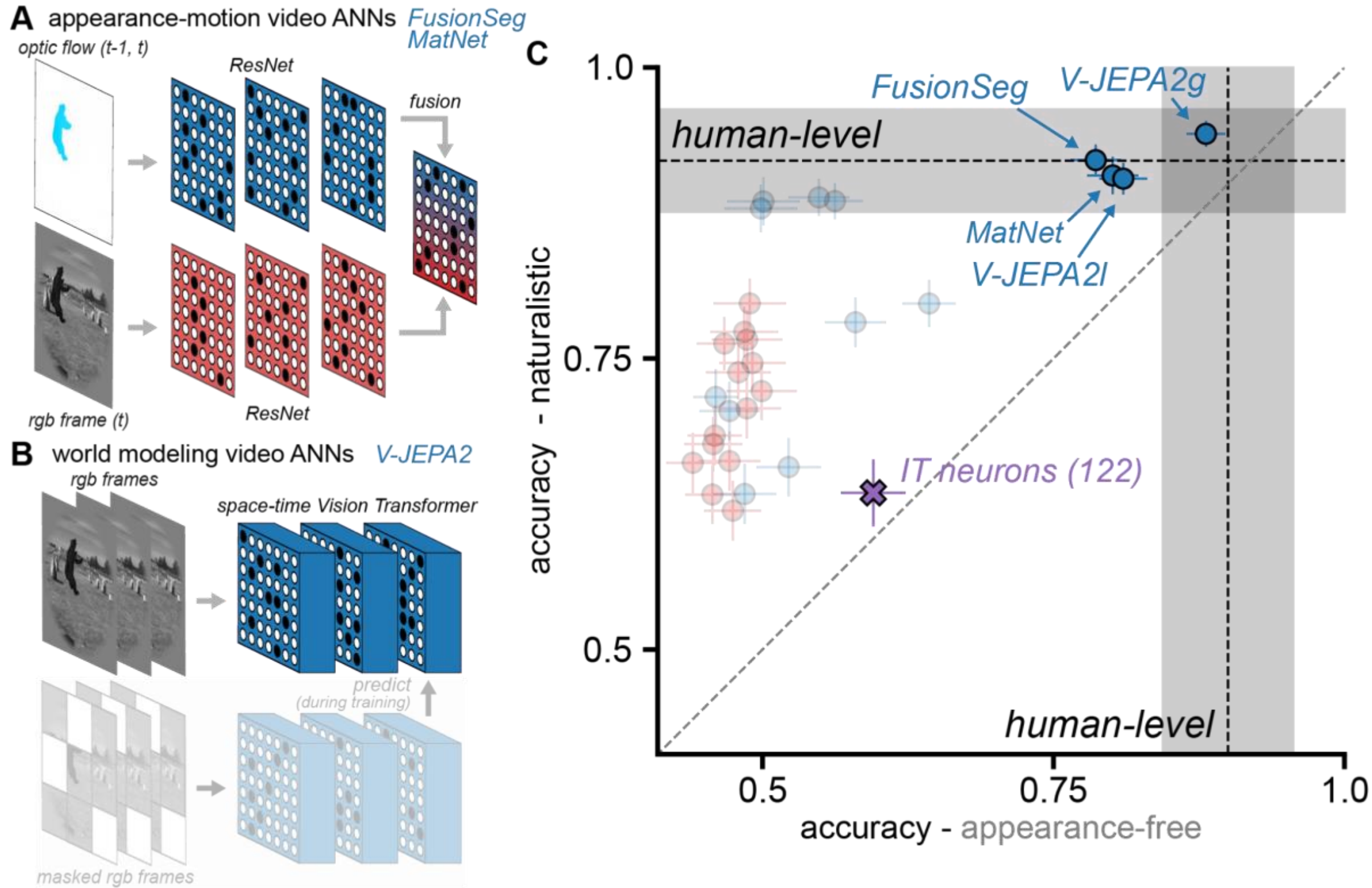

**Figure 4. Optic-flow-based and predictive video world modeling ANNs support robust cross-appearance motion decoding. A-B.** Schematic illustration of the two classes of video-based ANN architectures that showed the strongest appearance-invariant object motion direction decoding performance. **A.** Optic flow-based models (e.g., FusionSeg and MatNet) explicitly process motion signals extracted as optic flow between consecutive video frames and combine them with appearance

information from RGB video frames. **B.** Video world models (e.g., V-JEPA2) learn spatiotemporal representations directly from video sequences by predicting masked or future visual content during training, thereby encouraging the encoding of spatiotemporal structure beyond the level of pixel appearance. **C.** Generalization of motion-direction decoding across naturalistic and appearance-free conditions. Each point corresponds to a model and is plotted according to decoding accuracy when trained and tested on naturalistic videos (y-axis) versus tested on appearance-free videos (x-axis). The point indicates the maximum frame-indexed mean accuracy across videos. Error bars indicate the standard error across videos at the selected frame. Models lying on the diagonal exhibit identical performance across conditions and therefore encode motion representations that generalize across appearance changes. The vertical and horizontal lines report mean human behavioral performance in the two conditions, with the gray areas indicating the respective standard deviation. The purple point indicates the maximum time-indexed mean accuracy across videos achieved by the linear decoder from IT responses. Error bars indicate the standard error across videos at the selected time bin.

## Macaque IT contains motion information that generalizes across appearance disruption

We next asked whether the cross-appearance robustness observed in human behavior is reflected in the high-level ventral visual cortex. Human and macaque object recognition behavior is closely aligned under static [4] and dynamic[7] viewing, and population activity in macaque inferior temporal (IT) cortex supports and predicts object-related behavioral judgements[1,7,23,28,29]. Because IT also contains information about object motion [7], it provides a biological reference for asking whether motion-direction information remains accessible when recognizable appearance is removed.

We analyzed responses from 122 IT sites recorded in two macaques viewing the same paired naturalistic and appearance-free stimuli used in the behavioral and model experiments. At each time bin, a motion-direction decoder was trained on IT responses to naturalistic videos and tested, without refitting, on held-out naturalistic responses and the paired appearance-free responses (**Fig. S2A**). The decoder achieved above-chance accuracy on both held-out naturalistic responses (mean accuracy = 0.61 ± 0.02 SE; permutation $p < 0.001$) and appearance-free responses (mean accuracy = 0.58 ± 0.02 SE; permutation $p < 0.001$; **Fig. 4C**). The two distributions were not significantly different (*$p = 0.27$, $t(99) = 1.10$, paired t-test*), and cross-appearance decoding increased with the number of IT sites (**Fig. S2B**). Thus, motion-direction information learned from naturalistic IT responses remained readable when recognizable object appearance was removed.

These results demonstrate that IT contains motion-direction information that generalizes across substantial appearance disruption, identifying an appearance-robust dynamic representation in high-level ventral visual cortex. This neural result provides a biological target for distinguishing among artificial strategies that can achieve similarly robust behavior.

## Temporal processing improves late IT alignment, with predictive world models performing best

Our previous results showed that video-based ANNs provide a closer account of human behavior in dynamic object perception tasks based on naturalistic visual input. The high performance achieved by these developments suggest that machine learning-based spatiotemporal integration could be a critical computational mechanism for building dynamic visual representations, and may underlie the improved alignment of video-based ANNs with human behavior.

Based on this intuition, we next assess if the representations built through these operations can be used to explain the dynamics of neural activity in the IT cortex. To this end, we used noise-normalized Centered Kernel Alignment (CKA)[30] to quantify the similarity between ANN features and IT responses recorded over 200 videos with moving objects (20 videos per object, 10 object categories, 8 motion directions). Features from all video ANNs (N = 15 models) were extracted per-frame, whereas the IT population consisted of 404 neurons across three monkeys. Specifically, we compared: early model features (averaged across the first 5 video frames) with early IT responses (averaged between 90–180 ms), a time window known to contain object recognition relevant representations[3,23,24]; and late model features (averaged across the last 5 video frames) with late IT responses (averaged between 300–390 ms) (**Fig. 5A**), a time window known to contain object motion-related representations [7]. In this analysis, we contrasted the representation extracted from original video ANNs (spatiotemporal video ANNs, **Fig. 5B-C**, y-axis) with the representations from modified versions that do not perform any temporal integration but instead predict frame-specific features based solely on the visual appearance of each frame (spatial video ANNs, **Fig. 5B-C**, x-axis). Rather than comparing to standard feedforward image-based ANN models, this approach allowed us to isolate the specific contribution of temporal integration while holding architecture, training objective, parameter count, and representational capacity constant. This matched temporal-input manipulation allowed us to estimate the contribution of temporally varying information while preserving architecture and learned parameters.

We found that the early representations from frame-based and time-integrative video ANNs achieved similar CKA scores when compared with early IT responses (spatial video ANNs: *mean CKA = 0.35 ± 0.06 SD*; spatiotemporal video ANNs: *mean CKA = 0.34 ± 0.06 SD*, *p = 0.77, t(14) = 0.30, paired t-test,* **Fig. 5B**), suggesting that temporally varying input provided little additional benefit for explaining early IT representations. In contrast, when compared with late IT responses, spatiotemporal representations (*mean CKA = 0.30 ± 0.05 SD*; **Fig. 5C**) yielded significantly higher CKA scores (*p = 0.007, t(14) = -3.14, paired t-test*) than spatial features (*mean CKA = 0.28 ± 0.04 SD*; **Fig. 5C**). This result partially confirms that integrating information across frames provides a computational account of the dynamic evolution of IT neural responses. However, the improved CKA scores for spatiotemporal video ANNs do not imply full mechanistic alignment with IT. Notably, even the best spatiotemporal models achieve lower CKA values for late IT responses (**Fig. 5C**) than those obtained for early IT responses (**Fig. 5B**). Thus, although spatiotemporal integration increases correspondence with late-stage IT activity, current video ANNs still fail to fully capture the neural computations underlying dynamic object representations in IT.

At the level of specific modeling approaches, we find important differences. Models that explicitly integrate optic flow information (e.g., MatNet, FusionSeg) achieve robust, human-like performance on dynamic behavioral tasks, yet they do not explain late IT dynamics better than conventional optic flow-free convolutional representations (spatial: *mean CKA = 0.30 ± 0.01 SD*, spatiotemporal: *mean CKA = 0.28 ± 0.02 SD,* **Supp. Fig. S3C-D**). Video recognition models provide a modest improvement in explaining late IT responses (spatial: *mean CKA = 0.27 ± 0.05 SD*, spatiotemporal: *mean CKA = 0.29 ± 0.06 SD*). The strongest correspondence with IT emerges from recent large-scale-trained predictive world modeling architectures which consistently achieve the highest CKA scores (spatial: *mean CKA = 0.30 ± 0.02 SD*, spatiotemporal: *mean CKA = 0.36 ± 0.0 SD*) while also exhibiting robust dynamic object recognition behavior. These results suggest that large-scale video world modeling mechanisms may capture aspects of temporal processing in IT that are not accounted for by small-scale spatiotemporal video recognition ANNs alone.

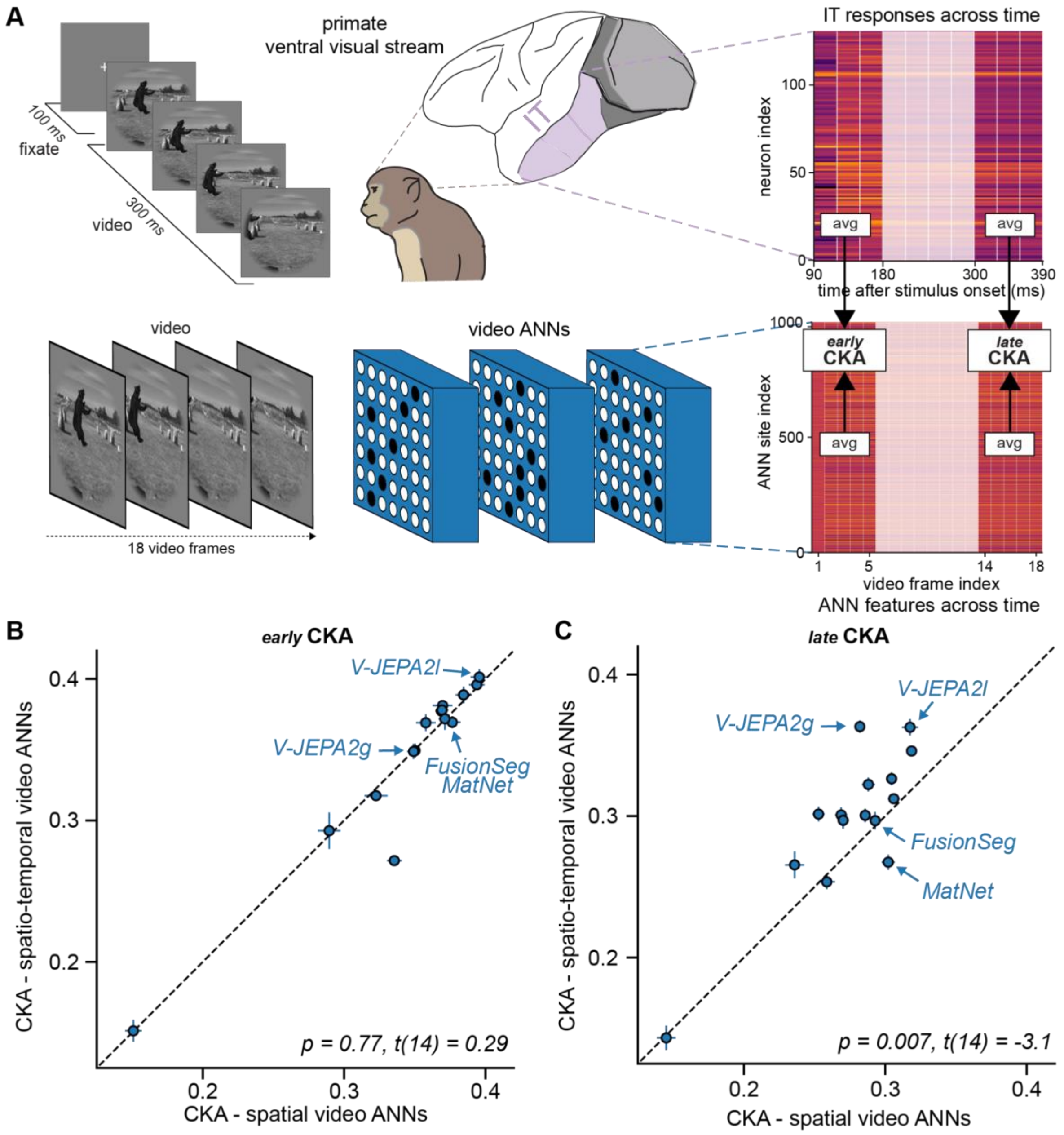

**Figure 5: Representational similarity between macaque IT responses and video ANN features during dynamic object viewing. A.** Schematic of the representational similarity analysis. Neural activity was recorded from macaque inferior temporal (IT) cortex (404 sites) while animals passively viewed 300 ms videos (N = 200) of moving objects. Based on previous analyses showing the emergence of object identity information at early latencies and motion information at later latencies, neural responses were averaged across an early window (90–180 ms after stimulus onset) and a late window (300–390 ms). Corresponding frame-wise features were extracted from video ANNs for the same 18-frame video sequences. Early and late model representations were obtained by averaging features from the first and last five frames, respectively. Representational similarity between neural and model responses was quantified using Centered Kernel Alignment (CKA), comparing early IT responses with early model features and late IT responses with late model features. **B.** Correspondence between early IT responses and representations of spatial and spatiotemporal video ANNs. Points indicate the CKA value for one model processing single frames independently (spatial processing, x-axis) and integrating multiple video frames (spatiotemporal processing, y-axis). Error bars report standard deviation across analysis repetitions. **C.** Correspondence between late IT responses and representations of spatial and spatiotemporal video ANNs. The same descriptive statistics for (**B**) are used.

## Current video models miss the temporal transformation of IT representations

Predictive world models provided the strongest IT correspondence, yet substantial model-brain differences remained. We therefore asked whether the residual gap reflects not only what information is present in the representation, but also how that representation evolves as the video unfolds. Specifically, we compared the temporal evolution of stimulus-response profiles in individual IT neurons with that of model feature dimensions. Prior work showed that object-identity-decodable IT representations emerge early (approximately 90-180 ms), whereas object-motion-decodable information becomes prominent later (approximately 300-390 ms)[7]. This temporal progression raises a stronger requirement for dynamic models: beyond integrating multiple frames, their internal representations should undergo a transformation comparable to the one observed in IT. We therefore quantified how rapidly neural and model response profiles changed across successive time points. We designed a neuron-level metric of temporal change that quantifies how neural activity evolves over time. The goal of this metric is to capture the transition from encoding static visual properties (e.g., object identity) to dynamic properties (e.g., motion direction and speed). For each neuron, we quantified how rapidly its stimulus-response profile changed across successive time points by comparing its response pattern at time t with those at *t+30* ms and *t+60* ms (**Fig. 6A-C**). Larger values indicate greater temporal evolution of the stimulus-response profile, whereas smaller values indicate greater temporal stability. By tracking this measure over time over multiple videos (3 monkeys, 393 IT neurons, 200 videos), we obtain a profile of how neural representations transition from stable to dynamic states, providing a quantitative description of temporal evolution in IT responses. Using this metric, we find that IT neurons at early post-stimulus latencies (early change, ~120 ms, *median = 0.27 ± 0.006 SE*, **Fig. 6D**) exhibit high temporal change with their subsequent activity, whereas at later latencies (late change, ~330 ms, *median = 0.11 ± 0.007 SE*, **Fig. 6D**) their activity change becomes significantly less over time. These results suggest that IT neurons contributing to object identity behavior undergo a temporal reorganization of their responses, transitioning toward a more stable activity regime at later time points (Δ *median early-late = 0.16, $p < 0.001$, $t(392) = 16.83$, paired t-test*). This increased temporal stability reflects a representational state in which dynamic information, such as object motion, can be more reliably read out. Applying the same metric to the units in video-based ANNs (**Fig. 7A**) reveals a key difference from IT dynamics. While these models do exhibit changes in their representations over time, their activity tends to rapidly stabilize, rather than undergoing the progressive transformation observed in IT. Specifically, unit responses at early video frames remain highly correlated with responses at later video frames, and this high similarity persists throughout the stimulus (**Fig. 7A-B**). This indicates that, unlike IT neurons, video-based ANNs show limited temporal evolution, with representations that are comparatively stable over time. We quantified this effect across the considered population of video models. At early video frames, the temporal integration of all ANNs show significantly reduced temporal change compared to IT neurons (early change, *spatiotemporal ANNs: median = 0.06 ± 0.01 SE, $p < 0.001$, Wilcoxon rank-sum;* **Fig. 7C**). At later stages (late change), this difference decreases, primarily because activity in IT neurons themselves becomes more stable over time. Importantly, video ANN features exhibit slightly greater temporal change than their frame-based non-time-integrative counterparts (late change, *spatiotemporal ANNs: median = 0.04 ± 0.01 SE, spatial ANNs: median = 0.02 ± 0.0 SE, $p < 0.001$, $t(14) = 5.55$, paired t-test,* **Fig. 7D**), indicating that the spatiotemporal architectures do introduce more dynamic representations.

However, this increase remains insufficient to match the magnitude and structure of temporal evolution observed in IT neurons. This temporal mismatch provides one candidate source of the residual representational gap between current video models and IT: although ANNs benefit from temporal integration and achieve higher decoding accuracy, they fail to reproduce the progressive representational transformations characteristic of IT.

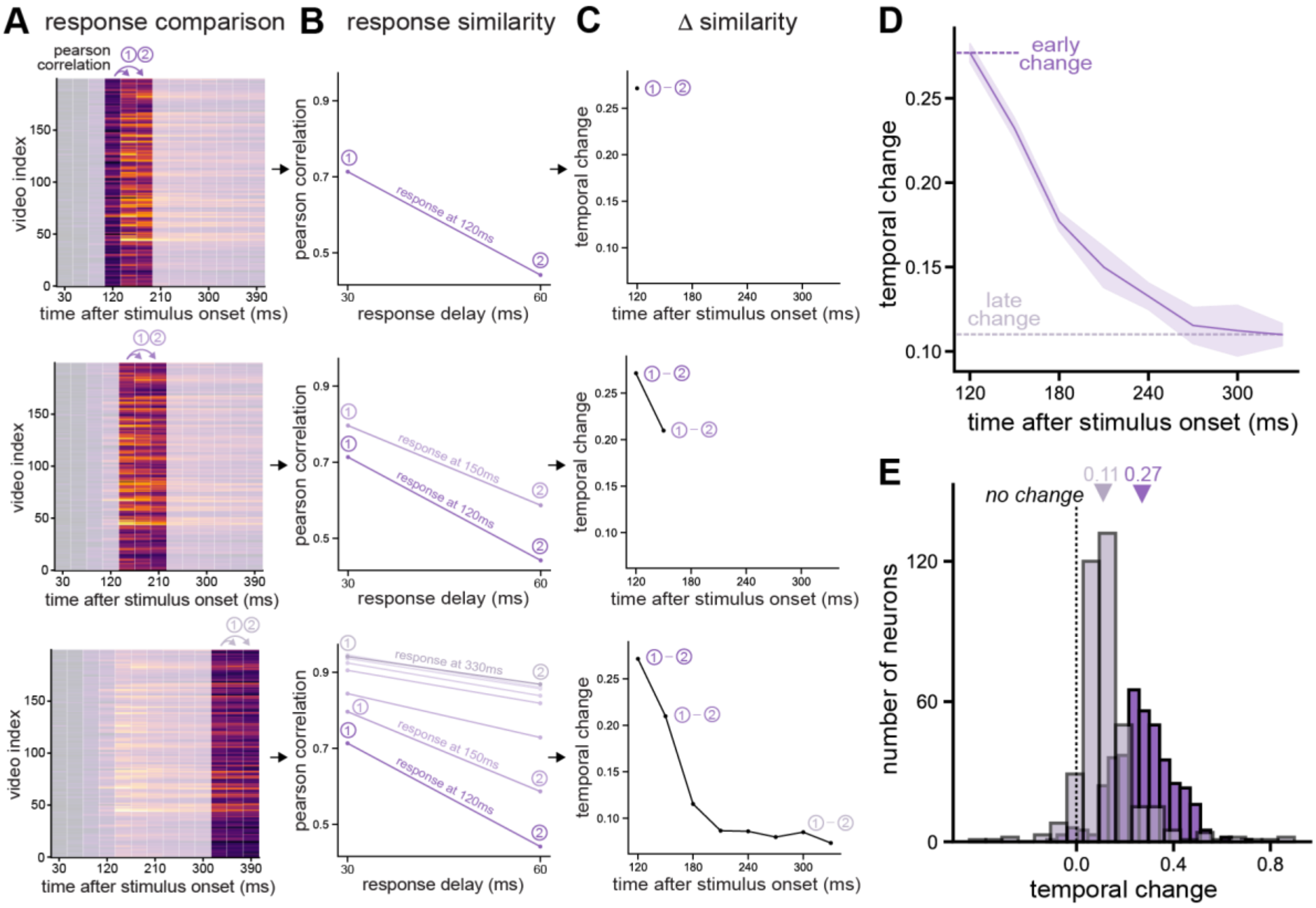

**Figure 6: Temporal evolution of neural responses in the inferior temporal (IT) cortex. A–C.** Schematic illustration of the temporal change metric computed at the single-neuron level. For each neuron, we analyzed its trial-averaged responses across time to a set of videos. At a given time point *t*, we computed the noise-corrected Pearson correlation between the response pattern at t and the responses at subsequent time points (*t+30* ms and *t+60* ms) across stimuli (**A**). These correlations quantify the temporal stability of the neural representation, with higher values indicating more similar responses across time (**B**). We then defined *temporal change* as the difference between the 30 ms-delay and 60 ms-delay correlations, capturing how rapidly the neural response evolves over time (**C**). Larger values indicate faster representational change, whereas smaller values indicate more stable activity. **D.** Time course of temporal change summarized across the IT population. The solid line indicates median while the shaded area indicates standard error, both over neurons. **E.** Distribution of temporal change values across neurons, comparing early and late response periods. Triangles indicate median values at the selected temporal bins.

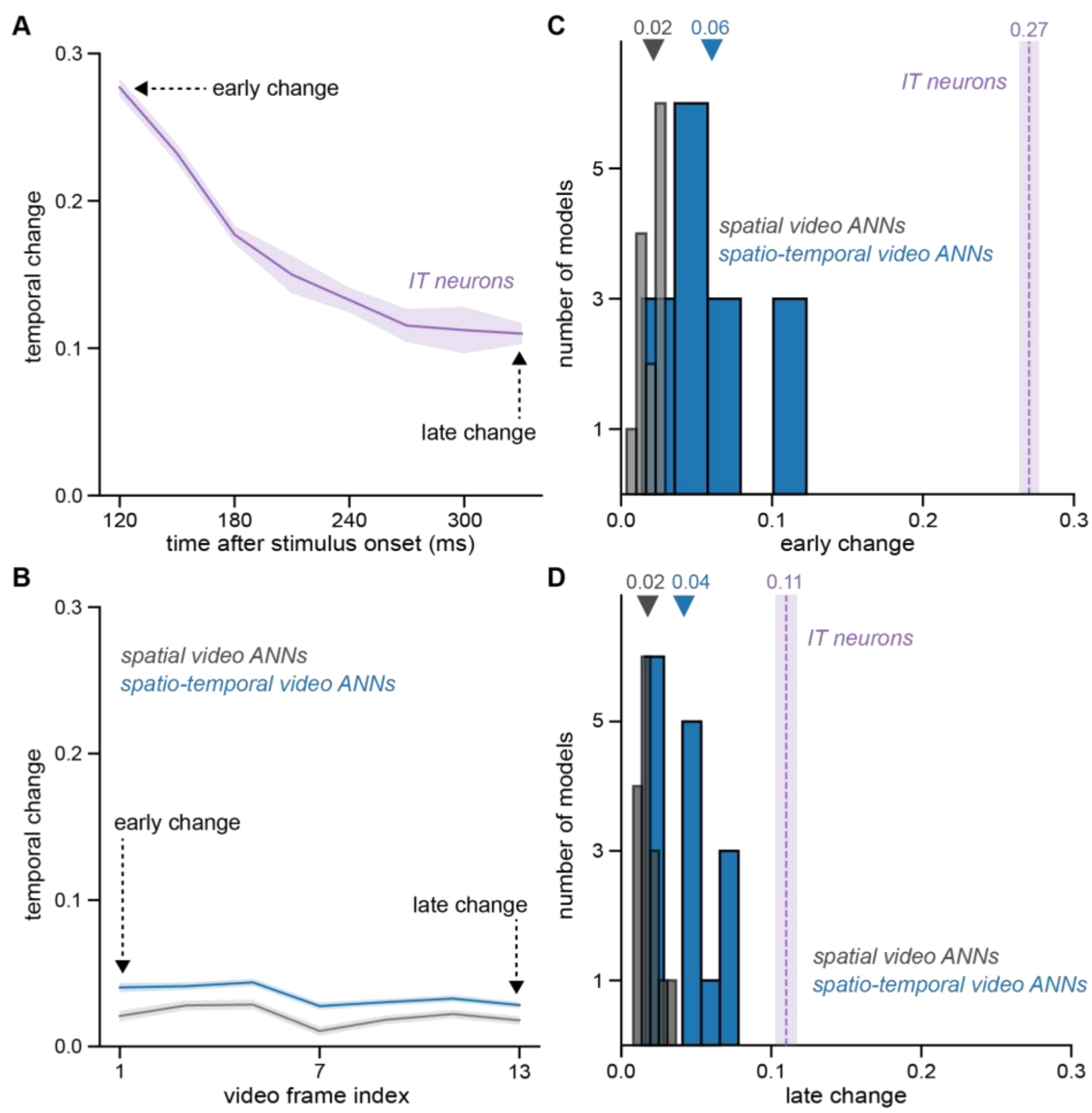

**Figure 7: Comparing temporal change in IT neurons and video ANN features. A-B.** Time-resolved analysis. **A.** The graph plots the time-course of temporal change averaged across IT neurons as a function of time after stimulus onset. The solid line indicates median while the shaded area indicates standard error, both over neurons. Arrows indicate the timepoints chosen for early (120 ms) and late (300 ms) change analysis. **B.** The graph plots the time-course of average temporal change of model sites (N = 1000 units) for frame-based video ANNs (gray) and time-integrative video ANNs (blue) across video frames. The solid line indicates median while the shaded area indicates standard deviation, both over units. Arrows indicate frame indices chosen for early and late change. **C.** Comparison of model distributions based on *early temporal change*. Histograms show the medians of units from individual models of each type, spatial frame-based video ANNs (gray) and spatiotemporal time-integrative video ANNs (blue). The purple region and dotted vertical line represent the standard error and median early change, respectively, for IT neurons. Specific group medians are marked above: IT neurons (purple line), time-integrative ANNs (blue triangle), and frame-based ANNs (gray triangle). **D.** Same comparison as in (**C**) but based on *late temporal change*. Histograms and regional markers for late change show a general shift to lower values for all groups, describing stability of the representations.

## IT progressively develops appearance-robust motion representations that remain weak in video models

The temporal-change analysis established that IT and current video models differ in how strongly their representations evolve, but it did not reveal what information is being reorganized. The cross-appearance results suggested a specific possibility: the critical transformation may involve how motion information is incorporated relative to object appearance. We therefore asked how strongly IT and model response patterns are preserved when either appearance or motion structure is selectively disrupted (**Fig. 8A**).

Motivated by recent work disentangling appearance and motion factors in video models[31,32], we asked whether individual IT neurons could be characterized by their sensitivity to appearance-based versus motion-based stimulus features. To this end, we designed a neuron-level metric based on the correlation of responses across controlled stimulus conditions. Specifically, we measured the firing rates of the same IT neuron in response to: (i) videos (N = 100) with coherent motion and naturalistic appearance (**Fig. 8A**), (ii) videos (N = 100) with the same appearance but incoherent motion, and (iii) appearance-free videos (N = 100) preserving the same motion. We quantified appearance-preserving similarity as the correlation between coherent and same-appearance/incoherent-motion responses, and motion-preserving similarity as the correlation between coherent and same-motion/appearance-free responses. Comparing conditions (i) and (ii) isolates appearance-driven responses independent of object motion, whereas comparing (i) and (iii) isolates motion-driven responses independent of appearance. For each neuron, we quantified appearance sensitivity as the noise-corrected Spearman correlation between responses to (i) and (ii), and motion sensitivity as the noise-corrected correlation between responses to (i) and (iii).

**Fig. 8B** (left) shows the time-resolved correlations of neural responses associated with appearance-based and motion-based stimulus factors. IT neurons (N = 122) exhibit high correlations between stimuli sharing the same appearance, indicating a strong and sustained sensitivity to structured object features. At later time points (~300 ms), we also observe significant correlations between stimuli sharing the same motion but differing in appearance (*$p = 0.02$, $t(81) = 2.39$, paired t-test*). This result indicates that, over time, IT neurons increasingly respond to motion-specific information that is invariant to appearance.

We applied the same analysis to the units of spatiotemporal video ANNs and their spatial counterparts to directly compare their representational structure with that of IT (**Fig. 8B**, right). Unlike IT neurons, whose responses evolve over time, ANN features remain largely stable, with correlations associated with appearance and motion factors showing minimal change across the stimulus. At early time points (~120 ms for IT; second video frame for ANNs), both systems behave similarly: IT neurons and ANN features predominantly encode appearance information, with no significant difference between spatial video ANNs and IT (*IT $\Delta$ median = 0.85 ± 0.06 MAD; spatial video ANNs $\Delta$ = 0.86 ± 0.06 SD*; *Wilcoxon rank-sum $p = 0.15$;* **Fig. 8C**). This indicates that time-independent model units capture the initial, appearance-driven stage of IT neurons' processing. However, clear differences emerge at later time points (~300 ms, second to last video frame). Consistent with the previous result, IT neurons develop computations that partially encode motion information independently of appearance, reflected in a reduced difference between appearance and motion correlations (*late $\Delta$ median = 0.34 ± 0.05 MAD;* **Fig. 8D**). In contrast, this transition is largely absent in ANN features: video ANN representations continue to be dominated by appearance-based units, showing much larger appearance–motion differences (*video ANNs $\Delta$ = 0.70 ± 0.16 SD;* **Fig. 8D**). On comparing temporal integration, at later video frames, we find that spatiotemporal video ANNs exhibit a smaller significant gap between appearance-based and motion factors than their spatial counterparts (*$p < 0.001$, $t(14) = -4.89$, paired t-test*) and image-based ANNs (**Fig. S4**), suggesting that spatiotemporal ANN architectures partially promote motion sensitivity. However, this effect remains insufficient to match IT dynamics.

The critical model-brain difference therefore emerges over time rather than at the initial response: both systems begin with strong dependence on appearance, but IT progressively incorporates motion structure in a form that generalizes across appearance, whereas this transformation remains weak in current video models. Together, these analyses identify the progressive incorporation of appearance-robust motion information into high-level object representations as a defining feature of primate dynamic vision - one that temporal integration alone does not reproduce. The earlier model comparisons further point to predictive world modeling as a promising computational route toward this biological solution, because it most strongly combined cross-appearance robustness with IT correspondence among the artificial systems tested.

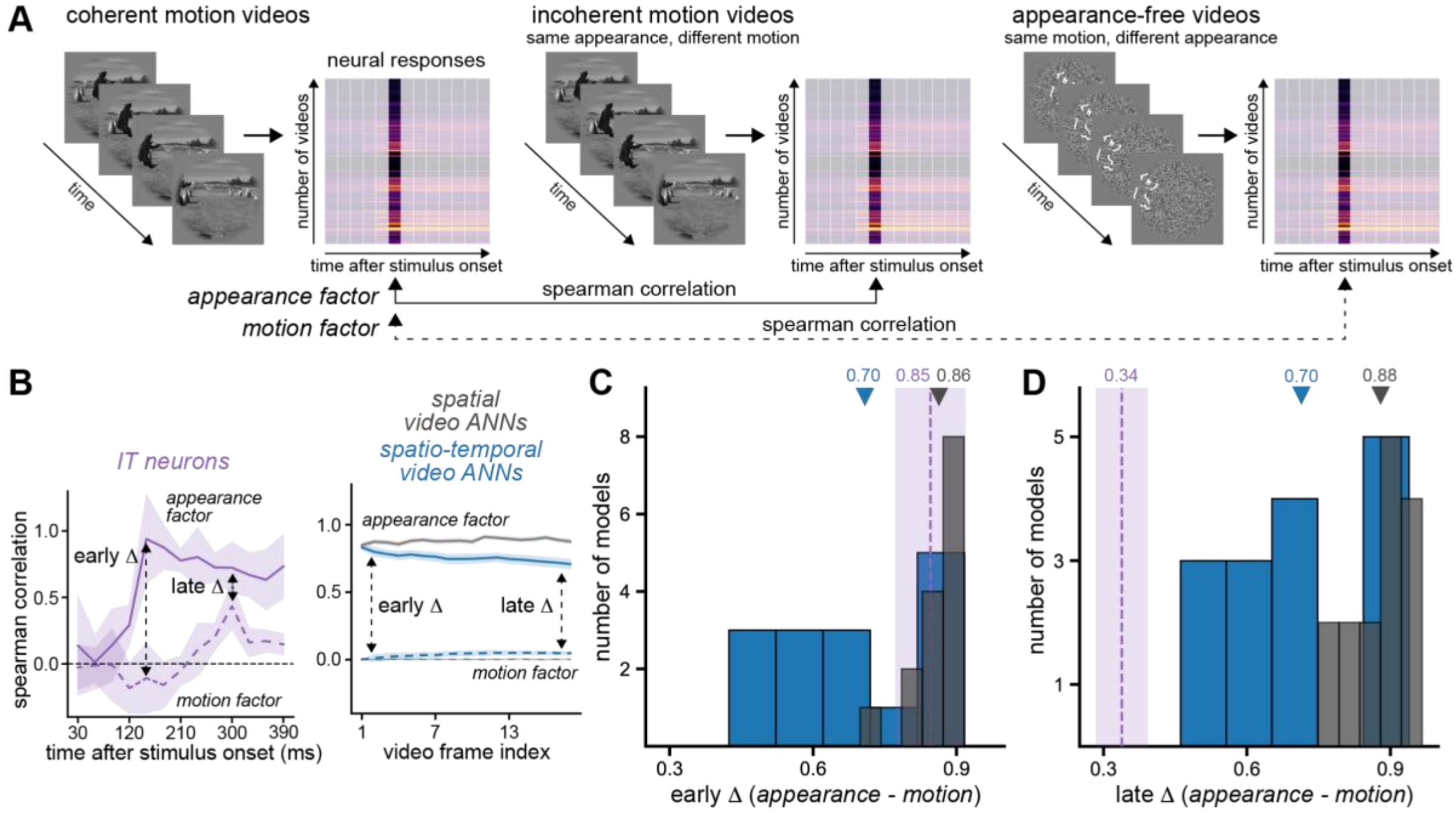

**Figure 8: Disentangling appearance and motion factors in IT neurons and video ANN representations. A.** Experimental design and metric. Neural and model responses were measured under three stimulus conditions: (i) naturalistic videos with coherent motion (left), (ii) videos with preserved naturalistic appearance but temporally incoherent motion (center), and (iii) appearance-free videos preserving motion but removing object appearance (right). For each neuron or model unit, we quantified appearance-driven and motion-driven responses by computing Spearman correlations of response patterns across stimuli: correlations between (i) and (ii) capture sensitivity to appearance, whereas correlations between (i) and (iii) capture sensitivity to motion independent of appearance. **B.** Time-resolved evolution of appearance and motion factors. Left: IT population responses. At each time point, median and standard error over neurons are reported. Right: spatial and spatiotemporal video ANN features. At each video frame, median and median absolute deviation over units are reported. **C–D.** Distribution across models of the difference between appearance and motion factors (appearance Spearman − motion Spearman), comparing non-time-integrative spatial video ANNs (gray), time-integrative spatio-temporal video ANNs (blue), and IT neurons (purple). **C.** Early response period: the purple dashed line and shaded area report median and median absolute deviation over neurons, respectively. For ANN models, the histogram of medians over units is shown. Specific group medians are marked above. **D.** Late response period: the same descriptive statistics used for (**C**) are reported.

# Discussion

Understanding how the brain represents a dynamic visual world remains a key challenge for both neuroscience and artificial intelligence. While deep neural networks trained on static images have become powerful models of ventral visual stream representations[2,3,33–35], the emergence of video-based architectures[10–13,16] has raised the possibility that these models may also capture the computations underlying dynamic visual perception[9,20,36,37]. Our results refine this view. Temporal integration clearly improves the accessibility of object identity, motion direction and speed from artificial representations, but temporal integration alone does not produce the robustness or representational dynamics observed in biological vision. Instead, our findings distinguish two computational requirements that are often conflated: integrating information across time and transforming that information into representations of object dynamics that generalize across changes in appearance.

This distinction becomes apparent when motion is dissociated from recognizable object appearance. Human observers remain highly accurate when object appearance is removed while motion structure is preserved, demonstrating that human dynamic perception relies on motion representations that generalize beyond specific visual features. Earliest popular ANN approaches to video recognition[10–12,15,25], in contrast, show strong performance on naturalistic videos but a substantial drop under appearance-free conditions (**Fig. 3**). Thus, successful motion decoding under naturalistic conditions does not necessarily imply that a model has acquired an appearance-robust representation of motion. Natural videos contain correlations between what objects look like and how they move, allowing systems to extract motion-related information without necessarily representing the underlying dynamic structure in a form that survives changes in appearance. The appearance-free manipulation therefore exposes a distinction that conventional benchmarks can obscure: a representation may contain substantial information about motion while still being strongly tied to the visual appearance from which that information was derived.

Importantly, this failure was not shared equally by all video ANNs. Optic flow-based models and predictive world modeling ANNs showed substantially stronger generalization across naturalistic and appearance-free conditions, approaching human-level performance (**Fig. 4**). These results suggest that robust dynamic representations are not an inevitable consequence of temporal processing; the computational strategy used to organize information across time matters. Architectures that explicitly encode motion through optic flow[18,26], or that learn predictive spatiotemporal structure through world modeling[17], are better able to support motion representations that generalize beyond appearance. In contrast, modern video object tracking ANNs[13] did not show comparable transfer, suggesting that large-scale video segmentation and object propagation mechanisms can remain tightly linked to appearance cues when they are not explicitly optimized to extract appearance-invariant motion information.

Behavioral robustness alone, however, does not establish that these different computational strategies converge on the same solution. The neural comparisons reveal a critical dissociation. Optic-flow-based models could support strong cross-appearance motion decoding while showing relatively limited correspondence with late IT representations. Predictive world models, by contrast, combined strong behavioral generalization with the highest correspondence to IT among the artificial systems tested. This result suggests that explicitly extracting motion can solve an important behavioral component of the task without reproducing the representational organization of the primate ventral stream. Predictive learning appears to provide a closer approximation to that biological solution, potentially because predicting latent visual structure requires a representation that preserves information about how objects evolve through time while remaining less dependent on their instantaneous appearance.

The neural data further identify what that biological solution entails. Activity in the macaque inferior temporal (IT) cortex contains reliable information about motion direction and support decoding that generalizes across appearance manipulations (**Fig. 4**). Although IT is traditionally viewed as a region specialized for object identity and form[23,28,29], these findings indicate that it also participates in dynamic visual processing at the object level[7]. This mirrors the robustness observed in human behavior and indicates that high-level ventral visual representations include object-centered dynamic information. Thus, the IT cortex appears to participate not only in recognizing what an object is, but also in representing how it moves in a way that is partially independent of its visual appearance. Biological dynamic vision therefore does not appear to discard appearance in favor of motion. Rather, high-level object representations evolve so that motion becomes integrated with object information while becoming progressively less dependent on the particular appearance through which that motion was observed.

Comparisons between IT responses and model representations further clarify the role and limits of spatiotemporal integration (**Fig. 5**). Video ANNs showed better correspondence with late IT activity than their spatial, non-time-integrative counterparts, indicating that temporal processing captures an important component of IT dynamics. However, this improvement was modest, and even the best models did not fully explain IT population representations. Video world modeling ANNs provided the strongest alignment with IT, consistent with their strong behavioral generalization, but substantial representational differences remained.

Temporal-response analyses provided further insight into why this gap persists. IT neurons exhibited a structured temporal evolution (**Fig. 6**): early responses changed rapidly, consistent with the emergence of visual and object identity-related information, whereas later responses became more stable, supporting the readout of dynamic variables such as motion direction. Video ANN features, by contrast, showed much weaker temporal reorganization (**Fig. 7**). Although spatiotemporal architectures introduced more dynamic activity than frame-based counterparts, they did not reproduce the progressive transformation observed in IT. This suggests that current models integrate information over time, but do not fully capture the temporal computations by which IT representations evolve from appearance-dominated responses toward dynamic object representations.

A related discrepancy emerged in the balance between appearance and motion coding (**Fig. 8**). Early responses in both IT neurons and ANN features were dominated by appearance, consistent with rapid object processing. Over time, however, IT neurons developed motion-related responses that generalized across changes in appearance. ANN features remained more strongly dominated by appearance, even in time-integrative models. This provides a mechanistic explanation for the behavioral dissociation: standard video ANNs improve performance through temporal integration, but they lack a representational stage in which motion becomes sufficiently disentangled from appearance.

Together, these findings nominate two linked principles for the development of more robust dynamic AI. The first is representational: dynamic object representations should progressively incorporate motion information in a form that generalizes across variation in appearance, while preserving the appearance information necessary for object recognition. The second concerns learning: predictive world modeling emerged as the strongest candidate among the strategies tested for approaching this biological organization. We interpret the latter cautiously. The predictive models examined here also differ from other architectures in scale, training data and implementation, so the present comparisons do not establish that predictive learning alone causes their stronger biological alignment. Nevertheless, the convergence of behavioral robustness and neural correspondence in these models motivates controlled tests of predictive objectives as a route toward learning more biologically faithful dynamic representations. Video world models and optic flow-based architectures provide promising future steps, but the remaining gap with IT suggests that current models still miss key neural computations.

More broadly, these results illustrate why model evaluation under naturalistic performance benchmarks alone can obscure meaningful computational differences between artificial and biological systems. If multiple stimulus dimensions covary, a model may achieve strong task performance by exploiting one source of information while failing to acquire another representation that is critical for biological robustness. Controlled manipulations that preserve one visual factor while disrupting another therefore provide a complementary benchmark for dynamic intelligence. Combining such interventions with neural measurements makes it possible to distinguish models that merely reproduce behavior from models that approach the representational strategies through which biological systems achieve that behavior.

Several limitations define important directions for future work. The present analyses focused on object identity, motion direction, and speed in controlled videos. Other dynamic properties, such as deformation, trajectory prediction, causal interactions, and multi-object dynamics, remain to be examined. In addition, the IT cortex is only one component of the broader dynamic vision system, which also includes dorsal stream regions such as MT[38,39] and parietal cortex[40]. Future work should therefore investigate how IT interacts with these areas and whether models that combine object-centered ventral representations with explicit motion-sensitive pathways better account for biological dynamic perception.

Dynamic vision exposes a computational challenge that is not solved simply by giving image-recognition systems access to more frames. Robust biological vision progressively transforms high-level object representations so that motion becomes integrated in a form that generalizes across appearance. Predictive world models currently come closest to combining this behavioral robustness with the corresponding neural representation, but they still fail to reproduce the temporal transformation observed in IT. The primate visual system therefore provides not only a benchmark for dynamic AI, but a concrete computational target: artificial systems should learn representations that evolve with visual experience, integrate appearance with motion, and preserve object dynamics across the changing appearances of the visual world.

# Methods

## Ethics Approval

*Human Experiments:* The study was approved by the York University Ethics Review Committee (Human Participant Review Subcommittee). Participants were recruited via the Amazon Mechanical Turk platform. Consent was presented on the first page; only consenting participants proceeded. The compensation rate was $15 CAD/hour.

*Non human primate experiments:* All data were collected, and animal procedures were performed, in accordance with the NIH guidelines, the Massachusetts Institute of Technology Committee on Animal Care, and the guidelines of the Canadian Council on Animal Care on the use of laboratory animals and were also approved by the York University Animal Care Committee.

## Human subjects

We collected large-scale psychophysical data from 107 human participants recruited via Amazon Mechanical Turk (MTurk). Participants completed the experiments online and were compensated at a rate of $15 CAD per hour. All experimental procedures involving human subjects were approved by the York University Human Participants Review Subcommittee and conducted in accordance with institutional guidelines. Participants received no additional training prior to the task.

Data quality for the behavioral discrimination tasks was ensured by excluding participants whose accuracy fell below 0.6 (chance level = 0.5) during the first 30 trials of each session. In addition, we assessed the reliability of video-level behavioral metrics as a function of the number of repetitions per video and observed that reliability increased with additional trials, reaching high values (≈0.8) and approaching asymptote after approximately 60 repetitions.

## Non-human primates

The nonhuman subjects in this study were 5 adult male rhesus monkeys (*Macaca mulatta*). All experimental procedures and data collection were conducted in accordance with NIH guidelines, the Massachusetts Institute of Technology Committee on Animal Care, and the guidelines of the Canadian Council on Animal Care for the use of laboratory animals, and were approved by the York University Animal Care Committee.
For the analyses presented in **Fig. 3**, **Fig. 4**, and **Fig. 8**, monkey M1 and monkey M2 contributed 61 sites each. For the analyses presented in **Fig. 5**, **Fig. 6**, and **Fig. 7**, monkey M3 contributed 192 sites, and monkeys M4 and M5 contributed 288 sites each.

## Visual Stimuli

High-quality videos of single objects were generated using the free ray-tracing software POV-Ray (http://www.povray.org), following procedures similar to previous studies[3,4,7,24]. Each video frame consisted of a two-dimensional (2D) projection of a three-dimensional (3D) object model (purchased from Dosch Design and TurboSquid) rendered against a random background. Ten object categories were used: bear, elephant, face, apple, car, dog, chair, plane, bird, and zebra. Object variation was introduced by manipulating six viewing parameters, position (x and y), rotation (x, y, and z), and size, allowing us to sample identity-preserving transformations. All image frames were achromatic and rendered at a native resolution of 256 × 256 pixels. Following similar prior experimental protocols to study perception during fixation[3,4,7,23,24], videos were rendered to last 300 ms at 60 Hz, resulting in 18 frames.

*Generation of videos for object identity, and object motion direction and speed behavioral tasks*

We first rendered two-dimensional images of the objects and composited them onto uncorrelated backgrounds, as described above. Based on the selected object speed (expressed in degrees per second, derived from pixel displacement per second for 256 × 256 frames spanning 8° of visual angle) and motion direction (eight categories: top, left, bottom, right, top-left, bottom-left, top-right, bottom-right), objects were then translated by the appropriate number of pixels using custom MATLAB scripts, while the background remained stationary.

*Generation of appearance-free videos*

To generate a noise pattern consistent with the stimulus motion, we first created a random noise frame defined over pixel space $p$. This noise field was then advected over time using a flow field $v(p,t)$, estimated from the original video with an optic flow algorithm, where $t$ indexes the video frames. Because the flow field $v(p,t)$ contains subpixel displacements and therefore does not align exactly with the discrete pixel grid, the warped noise frames were resampled using nearest-neighbor interpolation to realign them to the underlying pixel lattice. Further details of this procedure are provided in Ilic et al.[19].

*Generation of coherent-incoherent motion videos*

Starting from naturalistic coherent-motion videos, we generated a paired incoherent-motion stimulus by randomly permuting the temporal order of its 18 frames. This manipulation preserved the exact set of individual frames, and therefore the object identity, background, and distribution of visual appearances, while disrupting the original temporal sequence. Consequently, the videos no longer contained a smooth object trajectory or consistent motion direction and speed, while retaining the same duration, frame rate, and overall visual content as their coherent-motion counterparts.

## Behavioral Tasks

*Human behavioral tasks*

Human observers performed perceptual estimation tasks using the aforementioned video stimuli via Amazon Mechanical Turk. Each trial began with a fixation period of 100 ms followed by a 300 ms presentation of a video clip depicting a moving object. After the stimulus presentation, participants were instructed to report object attributes using an interactive interface. Three behavioral tasks were performed:

- *Object recognition:* Participants clicked on the perceived object category in a two-way choice interface.
- *Object motion direction discrimination:* Participants clicked on an arrow indicating the perceived direction in a two-way choice interface.
- *Object motion speed discrimination:* Participants were shown two videos with objects moving at different speeds and asked to select the video with the fastest speed in a two-way choice interface.

Participants were given up to 1500 ms to complete each response.

Although the stimulus sets comprised ten object categories and eight possible motion directions, the behavioral task used a two-alternative forced-choice procedure. On each trial, participants selected which of two candidate categories or directions matched the category or motion direction shown in the video. Thus, the ten categories and eight directions defined the complete set of stimulus categories, whereas each individual trial required a binary decision. Accordingly, chance-level performance was 0.5 rather than 0.1 (for object category) or 0.125 (for object motion direction).

*Human behavioral metric*

Behavioral performance was quantified using the video-level discriminability metric I1[4], adapted to responses collected in the two-alternative forced-choice task. On each trial, participants selected between the ground-truth class and one distractor class. For each video–distractor pairing, pairwise accuracy was calculated as the proportion of responses in which participants selected the ground-truth class. These accuracies were then averaged across all distractor classes to obtain a single I1 score for each video. The overall I1 score was calculated by averaging across videos. Scores range from 0 to 1, with 0.5 representing chance performance. This metric summarizes how reliably participants distinguished the correct class from the complete set of competing alternatives while accounting for possible differences in difficulty among distractor classes.

## Non-human primate electrophysiology

*Surgical Procedures*

Rhesus macaques were surgically implanted with multiple 10x10 micro-electrode arrays (Utah arrays; Blackrock Microsystems) in the IT cortex using a head post in a sterile environment. Each array comprises 96 electrodes measuring 1.5 mm long and spaced 400 μm apart from neighboring electrodes. The electrodes were connected through a percutaneous connector, facilitating simultaneous recordings across all 96 channels per array. The arrays were positioned anterior to the posterior middle temporal sulcus and inferior to the superior temporal sulcus in both brain hemispheres. Array placement was guided by the visible sulcal patterns during surgery. The surgical and animal procedures adhered to the Massachusetts Institute of Technology Committee on Animal Care and National Institutes of Health guidelines.

*Eye Tracking*

Eye movements were monitored using an EyeLink 1000 video eye-tracking system (SR Research). Through operant conditioning, subjects were trained to fixate on a central 0.2° white circle within a ±2° fixation window. Each behavioural trial commenced with an eye calibration procedure, where the monkeys made saccades to spatial targets and maintained fixation for 500ms. Calibration was repeated if any drift was observed.

*Passive fixation task*

Videos were presented for various tasks (see descriptions below) on a 24-inch LCD monitor (1,920 × 1,080 at 60 Hz) positioned 42.5 cm in front of the animal. During the passive viewing task, monkeys initiated each trial by fixating on a central white circle (0.2° diameter) for 100 ms. A video stimulus was then presented for 300 ms, followed by a 100 ms uniform gray blank screen, delivery of a fluid reward, and an inter-trial interval of 500 ms before the onset of the next trial. Trials were aborted if gaze deviated by more than ±2° from the fixation point at any time during video presentation.

*Neural Recordings*

Band-pass filtered (0.1 Hz to 10 kHz) neural activity was recorded by continuously sampling at a rate of 20 kHz using an Intan Recording Controller (Intan Technologies, LLC). The arrays' placement allowed for the neural sampling on the posterior to anterior axis of the IT cortex, and a majority of the data was based on multiunit activity. However, their specific spatial locations were not factored into the analyses. This study considered each site a random sample from a pooled IT population.

For population-level analyses (e.g. behavioral decoding), we used all the recorded IT sites. For the model-neural representational alignment and the neuron-specific correlational analyses, we included only neural recording sites that exhibited a significant visual response and a video rank-order response reliability greater than 0.4. Because most neural metrics were corrected for site-specific noise estimates, this inclusion criterion was not critical for the results and was primarily adopted to reduce computational load by excluding highly noisy recordings.

The reliability of individual neural sites was assessed using a split-half internal consistency measure computed across stimulus repetitions. This metric quantifies the stability of each site's response profile across random subsets of trials and serves as an inclusion criterion for downstream analyses. Specifically, for each neural site, trials were repeatedly divided into two random, non-overlapping halves. Within each split, mean responses were computed across repetitions for each stimulus, and the correlation between the two split means was taken as the raw split-half reliability. To correct for finite sampling due to data halving, we applied the Spearman–Brown correction. This procedure was repeated across 100 random splits, and the final reliability estimate for each site was defined as the average of the corrected correlations.

## Neural Data Analysis and Statistics

*Prediction of behavior from neural data*

Cross-condition linear decoder transfer: Neural decoding analyses were performed independently at each 30-ms time bin. For each time bin ($t$), trial-averaged responses were organized into a population-response matrix (N videos x S neural sites). Object motion direction (**Fig. 4**) was decoded using Linear Discriminant Analysis (LDA), implemented in Scikit-learn, with eight output classes corresponding to the eight motion directions. Decoder performance was estimated using 10-fold cross-validation. To determine whether IT population responses encoded motion information that generalized across changes in object appearance (**Fig. 4** and **Supp. Fig. S1**), we performed cross-condition decoder transfer. For each 30-ms time bin and cross-validation fold, an LDA motion-direction decoder was trained exclusively on IT responses to the naturalistic training videos. Without refitting, the trained decoder was evaluated on neural responses to both the held-out naturalistic videos and their paired appearance-free counterparts. Naturalistic and appearance-free versions of the same video were always assigned to the same cross-validation fold, ensuring that paired stimuli did not occur in both the training and test sets. The temporal index leading to the highest naturalistic score was retained to select the corresponding appearance-free score. Decoding performance on the appearance-free responses therefore quantified the extent to which the motion-direction representation learned from naturalistic responses generalized when recognizable object appearance was removed.

Decoding metric: As for the human behavioral tasks, performance was quantified using the image-level discriminability metric (I1) adapted to videos. For each video, we compared the predicted probability assigned to the ground-truth class with the probability assigned to each alternative class. Each comparison was expressed as the probability of the correct class divided by the summed probabilities of the correct and alternative classes. These pairwise values were then averaged across all alternative classes to obtain a single I1 score for each video. The final score was calculated by averaging across videos, and variability was reported as the standard error of the mean.

*Temporal change of neurons*

To characterize the temporal stability and evolution of neural representations (**Fig. 6**), we analyzed trial-wise firing rates of individual IT neurons across successive time bins. For each neuron and for each time bin, firing rates were first averaged across trials to obtain a stimulus-wise response vector at each time point. Temporal similarity was then quantified by computing the Pearson correlation between the response vectors at $t$ and $t+30$ *ms* and $t+60$ *ms*.

To correct for noise and finite trial variability, we estimated the reliability of each neuron at each time point using a split-half correlation procedure. Trials were randomly divided into two halves, correlations were computed across stimuli, and the resulting values were corrected using the Spearman–Brown formula to estimate single-neuron reliability. Temporal correlations between time points were subsequently normalized by the geometric mean of the reliability estimates at the corresponding time bins, yielding noise-corrected measures of temporal consistency.

For population-level analyses, correlations were aggregated across repetitions using the median and then summarized across neurons. Analyses were restricted to early and late response windows (up to 390 ms post-stimulus onset). To quantify the rate at which neural representations change over time, we computed a temporal change measure defined as the difference between correlations at short ($t+30$ ms) and longer ($t+60$ ms) temporal offsets from a given response origin. To ensure robust estimation, an adaptive starting time was determined for each neuron based on its response reliability. Measures were computed only for neural sites exceeding a minimum reliability threshold of 0.4.

*Selectivity of neurons to appearance and motion factors*

To dissociate neural sensitivity to appearance-based and motion-based stimulus factors, we analyzed IT responses to three classes of videos (see **Fig. 8A**): (i) coherent-motion videos with naturalistic appearance; (ii) videos with naturalistic appearance but incoherent motion; (iii) and videos with identical coherent motion but no appearance information.

For each neuron and each time bin, we computed stimulus-wise response vectors by averaging firing rates across trials. We then quantified neuron-level similarity across stimulus conditions using rank-based (Spearman) correlations. Specifically, correlations were computed between responses to coherent and incoherent videos ((i) vs. (ii)) to capture sensitivity to appearance-based factors, and between responses to coherent and appearance-free videos ((i) vs. (iii)) to capture sensitivity to motion-based factors.

Correlations reflecting appearance and motion sensitivity were then corrected for noise by normalizing them by the geometric mean of the Spearman–Brown–corrected split-half reliabilities of the corresponding stimulus conditions. This procedure yielded noise-corrected estimates of the degree to which individual neurons encode motion-specific or appearance-specific information over time.

Time-resolved population statistics were obtained by computing the median corrected correlation across neurons at each time bin, with variability estimated as the median of absolute deviations across neurons. These trajectories were used to compare the evolution of appearance and motion encoding across early and late response phases. In addition, we examined the distribution of neuron-level correlations at representative early and late time points by plotting histograms of corrected correlations, allowing us to assess heterogeneity within the neural population. Only neurons passing split-half reliability > 0.0 were retained.

## ANN Model Analysis

*Model Types*

For video-based ANNs, we considered N=15 state-of-the-art generalistic methodologies developed for different video-based computer vision tasks[31,32]. Specifically, we used: MatNet[26], FusionSeg[18], SAM2[13], SlowR50[12], SlowR50-ssv2[12], I3D[11], X3D-XS[25], VideoMAE[41], C2D[42], VideoMamba[15], VideoMamba-ssv2[15], TimesFormer[10], TimesFormer-ssv2[10], V-JEPA2l[17], V-JEPA2g[17].

For image-based ANNs, we considered N=14 popular architectures benchmarked on BrainScore[2], specifically: Inception-V3[43], PNASNet[44], ConvRNN[45], NASNet[46], CORnet-S[47], EfficientNet-B0[48], DenseNet-121[49], AlexNet[35], ResNet-50[34], ResNet-50-SSL[34], ViT[50], ViT-SSL[50], ConvNeXt[51], Hiera[52].

In **Supp. Fig. S5**, we report more details on the implemented ANN architectures. For each architecture, we describe: name; the layer used as IT model; the task for which they were optimized; the neural network architecture category; and the dataset used for the training (IN: ImageNet, K400: Kinetics-400, SSv2: SomethingSomethingv2, DAVIS, SA-V, YT-T-1B: YT-Temporal-1B, H100M: Howto100M).

*Model Training*

All ANNs used in this study were employed off the shelf, using publicly available pre-trained models from their respective repositories. No additional training or fine-tuning was performed. For initialization-based video ANNs (e.g., SAM2[13]), the object bounding box in the first frame was manually annotated and provided as a prompt prior to the inference phase.

*Model Feature Extraction*

For image-based ANNs, we extracted frame-wise features from the IT-like layer specified in the Brain-Score platform. Recurrent models such as CORnet-S and ConvRNN were treated as feedforward image-based models by processing each video frame independently and extracting features from their reported IT-like layer. For video-based ANNs with a fixed temporal buffer of N input frames (e.g., video recognition models, video world models), features were extracted at each frame by populating the buffer with the current frame and the N−1 preceding frames in a First-In-First-Out manner. When the buffer was larger than the number of previous frames available, we filled it by repeating the first frame. For video-based ANNs that produce frame-wise outputs (e.g., SAM2, MatNet, FusionSeg), features were extracted directly during the computation of each frame. For optic flow-based models, optic flow fields between consecutive frames were computed with RAFT[27].

To identify the most IT-like layer for each video ANN, we extracted features from multiple, equally spaced layers using the images of the HVM640 dataset[53]. For each layer, we computed neural predictivity to predict macaque neural responses for the same images. The layer achieving the highest percentage of explained variance (%EV) was selected as the IT-like layer of the video ANN.

For the analyses on spatial non-time-integrative versions of video ANNs (**Fig. 5**, **7**, and **8**), we used the same models described in **Supp. Fig. S5B**, but disabled their temporal integration mechanisms. Specifically, we enforced purely spatial processing by removing any temporal signal from the inputs. For buffer-based models, this was achieved by feeding the same frame repeatedly to fill the temporal buffer while processing each frame. For optic flow-based models, we computed optic flow between identical copies of each frame, thereby eliminating motion signals. For SAM2, we initialized and ran the model on each video frame independently.

For all the considered models, feature representations were compressed to a population of 1000 features using Sparse Random Projection (SRP)[3]. 10 repetitions of SRP were used for all the model analyses.

*Prediction of behavior from model features*

Linear decoding: ANN representations were evaluated using linear decoding analyses performed independently at each video frame. For each frame (*t*), the feature matrix (N videos x 1000 features), was used to train a decoder implemented in Scikit-learn. Object motion direction was decoded using Linear Discriminant Analysis (LDA) with eight output classes corresponding to the eight motion directions. Object identity was decoded using the same procedure with ten output classes corresponding to the ten object categories. Decoder performance was estimated using 10-fold cross-validation. In each fold, the decoder was fitted exclusively on the training videos and evaluated on the held-out videos. Object speed was decoded using Partial Least Squares (PLS) regression, with the continuous motion speed provided as a scalar target. All decoding analyses were repeated for each of the ten Sparse Random Projection realizations, and results were aggregated across projection repetitions.

Cross-condition decoder transfer: To assess whether ANN representations encoded motion information that generalized beyond object appearance (**Figs. 3** and **4**), we performed cross-condition decoder transfer. For each video frame and cross-validation fold, an LDA motion-direction decoder was trained exclusively on representations of the naturalistic training videos. Without refitting or otherwise exposing the decoder to the target domain, it was then evaluated on representations from both the held-out naturalistic videos and their paired appearance-free counterparts. Naturalistic and appearance-free versions of the same video were always assigned to the same cross-validation fold, preventing paired stimuli from appearing in both the training and test sets. The frame index leading to the highest naturalistic score was retained to select the corresponding appearance-free score. Performance on the appearance-free condition therefore measured the extent to which motion-direction information learned from naturalistic videos transferred when object appearance was removed.

Decoding metric: To quantify the decoding accuracy of models in object recognition and motion direction, we used the same metric described for the decoding of behavior from IT neurons. Object-speed decoding performance was quantified using pairwise ranking accuracy. PLS regression was applied independently at each video frame to generate an out-of-sample scalar speed prediction for every video. At each frame, every video was compared with all other videos having a different ground-truth speed. A comparison was scored as correct when the ordering of the two predicted speeds matched the ordering of their ground-truth speeds and as incorrect otherwise. Pairwise scores were first averaged across all comparisons involving each video, excluding self-comparisons and equal-speed pairs, and were subsequently averaged across videos to obtain the overall decoding accuracy at that frame. Scores ranged from 0 to 1, with 0.5 indicating chance-level pairwise ordering performance. This procedure placed regression performance on the same interpretable scale as the two-alternative human speed-discrimination task.

*Model–neural representational similarity*

Representational similarity between ANN features and macaque IT responses was quantified using noise-normalized linear Centered Kernel Alignment (CKA)[30]. For the analysis in **Fig. 5**, model and neural responses were measured for the same 200 naturalistic videos. Early IT representations were obtained by averaging firing rates within 90–180 ms after stimulus onset, whereas late IT representations were averaged within 300–390 ms. Corresponding ANN representations were obtained by averaging features over the first five and last five video frames, respectively. For each model, units were compressed to 1000 feature dimensions using Sparse Random Projection, and CKA was computed independently for each of 10 projection repetitions. Feature and neural-response matrices were mean-centered across stimuli before calculating linear CKA. To account for trial-to-trial variability in the neural responses, IT reliability was estimated by repeatedly dividing stimulus repetitions into two non-overlapping halves, computing CKA between the resulting neural representations, and applying the Spearman–Brown correction. Raw model–IT CKA values were normalized by the square root of the corrected neural reliability; no reliability correction was applied to the ANN features. The same procedure was applied to the original spatiotemporal models and their spatial, non-time-integrative counterparts. CKA scores were summarized across projection repetitions, and paired comparisons between spatial and spatiotemporal representations were performed across models. For the analyses in **Supp. Fig. S3**, the procedure was repeated using responses from 122 IT sites to the 100 paired naturalistic and appearance-free videos.

*Temporal change of ANN representations*

To characterize the temporal stability and evolution of ANN representations, we applied the same analysis pipeline used for IT neurons to the units of each ANN. Two modifications were introduced. First, temporal correlations were computed between response vectors at frame *t* and frames *t+2* and *t+4*, corresponding to temporal offsets of approximately 30 ms, to match the temporal separation used in the neural data. Second, because ANN activations are deterministic, correlations were computed without applying noise-reliability corrections. Early and late network responses were defined as the activations at the first and the 14th frames of the video, respectively.

*Selectivity of neurons to appearance and motion factors*

To dissociate sensitivity to appearance-based and motion-based stimulus factors in ANN representations, we applied the same analysis pipeline used for IT neurons to the units of each network. The only modification was that Spearman correlations were computed without noise-reliability correction, as ANN activations of pretrained models are deterministic. Early and late network responses were defined as the activations at the second and second-to-last frames of the video, respectively.

## Statistical Analyses

All statistical analyses were performed in Python using the scipy.stats library. Relationships between continuous variables (e.g., human prediction, neural or model decodes) were assessed using Spearman rank or Pearson correlation coefficients, unless otherwise specified.

Group comparisons followed a consistent procedure based on the distributional properties and pairing of the data. Normality of each distribution was assessed using the Shapiro–Wilk test. When both groups satisfied the normality assumption, parametric tests were applied: a paired t-test for paired comparisons (e.g., within-subject or within-model analyses) and an independent-samples t-test for unpaired comparisons (e.g., between models). If the normality assumption was violated for at least one group, non-parametric tests were used instead. Specifically, we applied the Wilcoxon rank-sum test (equivalent to the Mann–Whitney U test) for unpaired comparisons. All tests were two-tailed, and exact p-values and test statistics are reported.

The analyses in this study involve several conceptually independent comparisons (e.g., model-to-human, model-to-brain, architecture-specific trends). Because these analyses test distinct hypotheses, each was evaluated within its own statistical family rather than pooled across the entire manuscript. Following common practice in computational and systems neuroscience, we therefore report exact p-values and test statistics without applying global multiple-comparison corrections. Results that do not reach statistical significance are described as non-significant trends and are not used to support inferential conclusions.

## Code Availability

Code to replicate the results is available at : https://github.com/vital-kolab/dyn-align/

# Supplementary Figures

Primate vision reveals a missing principle for robust dynamic AI

*Matteo Dunnhofer, Christian Micheloni, Kohitij Kar*

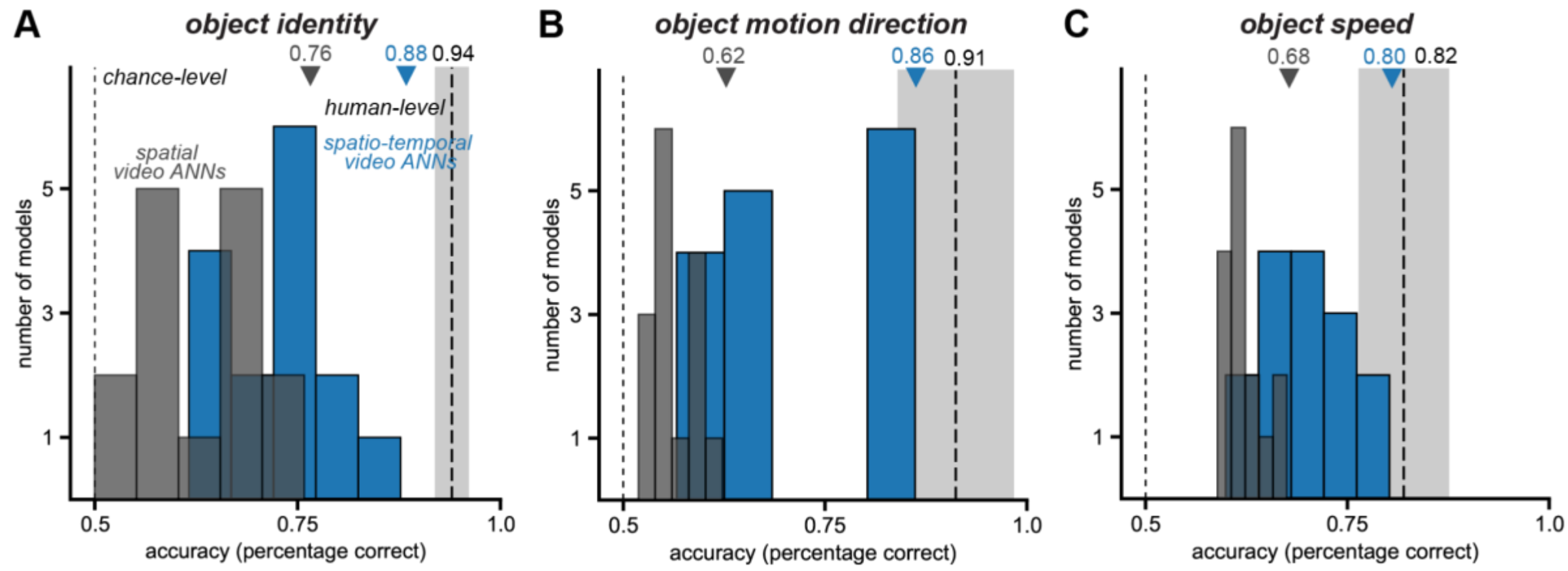

**Figure S1: Decoding behavioral variables from spatial versus spatiotemporal video ANN representations. A–C.** Distribution of decoding performance across all video models considered in this study (N = 15) for three behavioral variables: object identity, object motion direction, and object speed. Histograms show performance for spatial (non-time-integrative) video ANNs (gray) and spatiotemporal video ANNs (blue), with triangles indicating the maximum decoding accuracy achieved within each group. **A.** Object identity decoding. Spatiotemporal ANNs achieve higher peak performance (*mean accuracy = 0.72 ± 0.07 SD*) compared to spatial ANNs (*mean accuracy = 0.63 ± 0.07 SD*), indicating significantly improved extraction of appearance-based motion information when temporal context is available (Δ *mean accuracy = 0.09, p < 0.001, t(14) = 4.57, paired t-test*). **B.** Motion direction decoding. Spatiotemporal ANNs significantly outperform frame-based models (spatiotemporal video ANNs *mean accuracy = 0.71 ± 0.11 SD*; spatial video ANNs *mean accuracy = 0.56 ± 0.03 SD*; Δ *mean accuracy = 0.15, p < 0.001, t(14) = 5.93, paired t-test*), highlighting the importance of temporal integration for encoding motion direction. **C.** Object speed decoding. Similarly, spatiotemporal ANNs achieve significantly higher accuracy than frame-based ANNs (spatiotemporal video ANNs *mean accuracy = 0.70 ± 0.06 SD*; spatial video ANNs *mean accuracy = 0.62 ± 0.03 SD*; Δ *mean accuracy = 0.08, p < 0.001, t(14) = 4.27, paired t-test*), reflecting enhanced sensitivity to dynamic features. Across all tasks, spatiotemporal (video-based) architectures consistently outperform non-time-integrative spatial (frame-based) models, demonstrating that temporal processing improves the linear decodability of both static (identity) and dynamic (motion direction and speed) information from model representations.

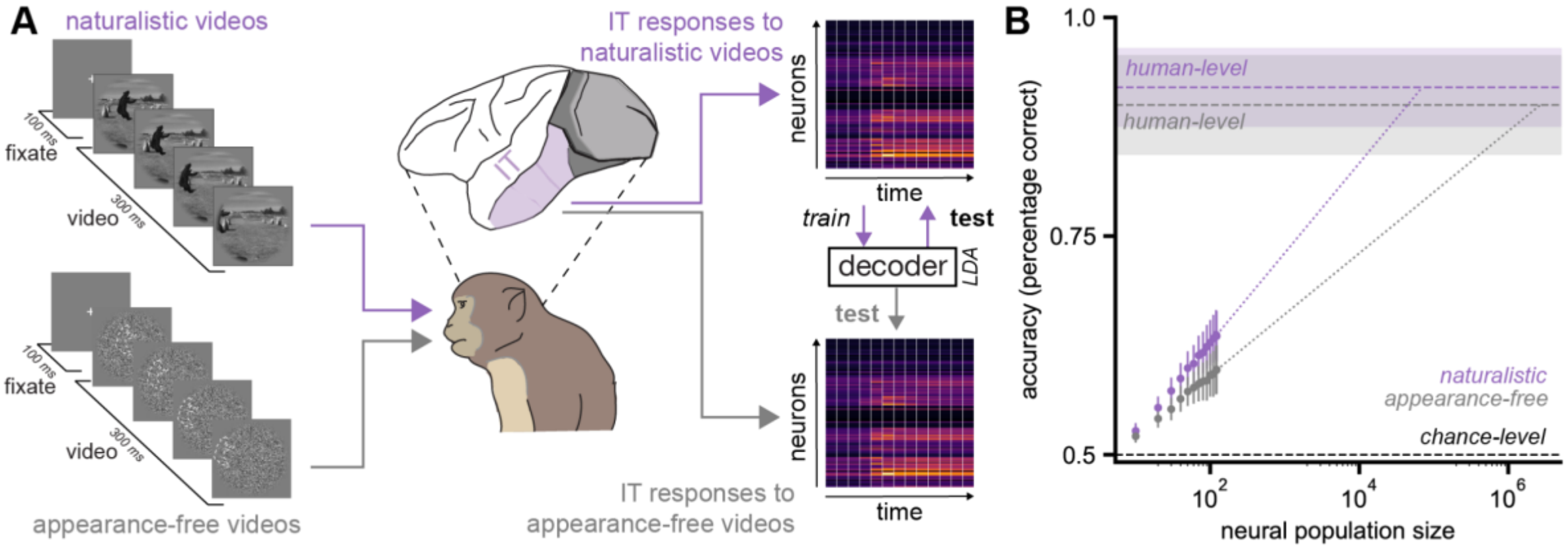

**Figure S2: Exposing appearance-dependent versus appearance-invariant motion representations in IT neurons. A.** Analysis pipeline. Neural responses were recorded from 122 sites in macaque inferior temporal (IT) cortex while animals passively viewed two stimulus conditions: naturalistic videos (100 videos), which contained both object appearance and motion, and appearance-free videos (100 videos), in which the same object motion was preserved but visual appearance was replaced with noise. A linear decoder (Linear Discriminant Analysis, LDA) was trained to predict object motion direction from IT responses to naturalistic videos and was then tested on responses to both held-out naturalistic videos and paired appearance-free videos. This cross-condition decoding procedure directly assesses whether IT motion representations generalize when object appearance is removed. **B.** Population scaling of IT decoding performance. Motion-direction decoding accuracy is shown as a function of the number of IT sites included in the decoder. Decoders were trained on responses to naturalistic videos and tested on naturalistic and appearance-free responses. Accuracy increased with population size in both conditions, indicating that IT population activity contains motion-direction information that generalizes across changes in object appearance. Points indicate the maximum time-indexed mean decoding accuracy across videos. Error bars indicate the standard error across videos at the selected time bin.

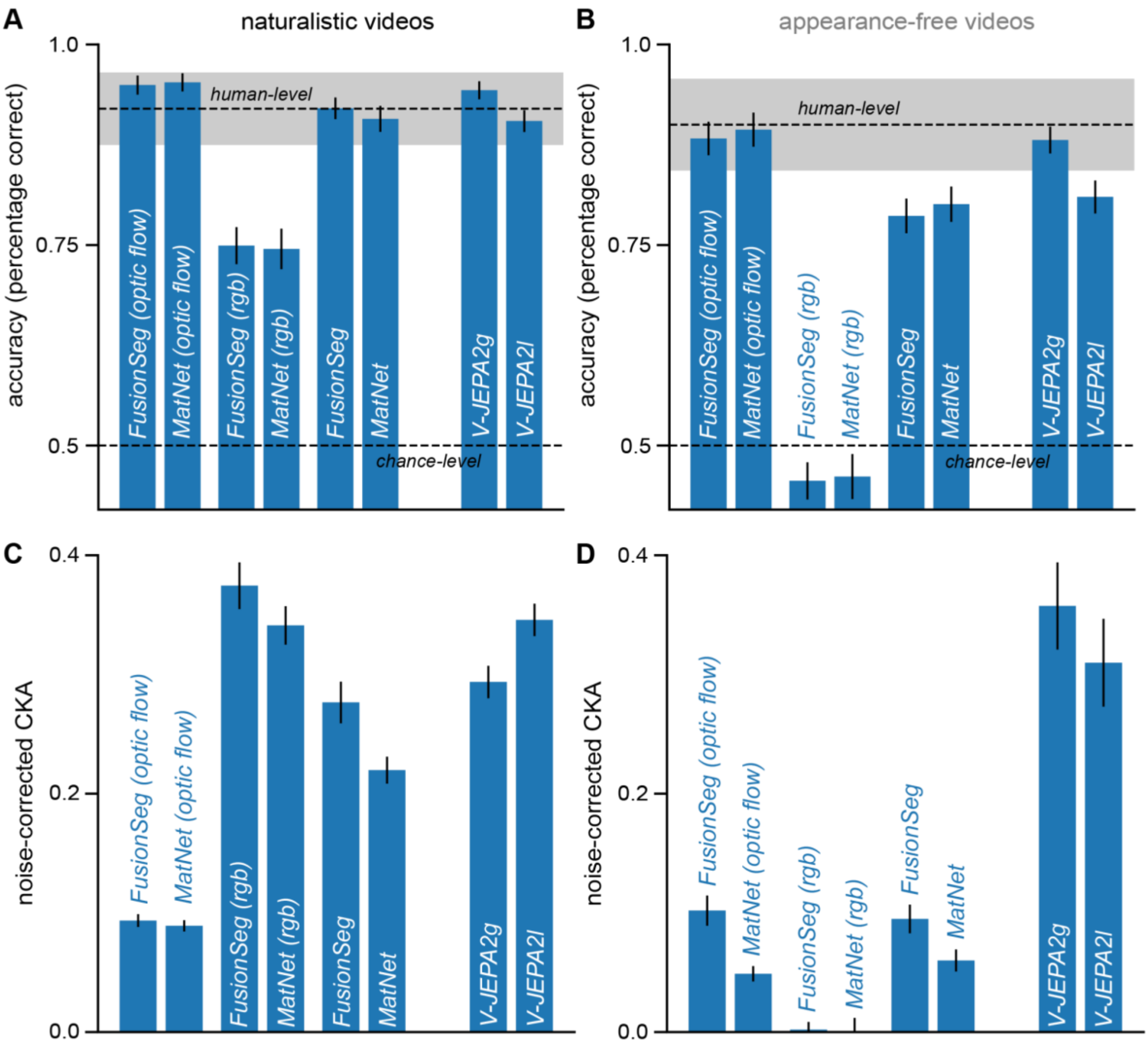

**Figure S3: Behavioral and neural alignment of optic flow-based and video world modeling ANNs. A.** Bars indicate maximum decoding accuracy from ANN representations over video frames. Error bars indicate standard error over videos. Optic flow-based models and video world models achieve high accuracy when object appearance is available. For optic flow-based methods (FusionSeg, MatNet), we also report the accuracy of decoding from their visual (rgb) and motion (optic flow) network stream representations (**Fig. 4A**). The dashed horizontal line indicates chance-level performance. **B.** Motion-direction decoding performance

on appearance-free videos. Under this condition, object motion is preserved but recognizable appearance is removed. Optic flow-based models maintain high decoding accuracy thanks to the motion branch, whereas appearance-only stream representations show a reduction to chance-level performance. Predictive world models maintain high accuracy even in appearance-free conditions. **C.** Model–IT response representational similarity for naturalistic stimuli. Bars show the correspondence between late ANN representations (last 5 video frames) and late IT population activity (300-390 ms window, 122 sites, 100 videos), quantified using CKA. Error bars report standard deviation over analysis repetitions. Video world models and appearance-based representations show moderate alignment with late IT responses. Optic flow-only representations show a clearly reduced alignment with late IT dynamics. **D.** Model–IT representational similarity for appearance-free stimuli. Video world models show the strongest correspondence with late IT motion-induced responses, whereas optic flow-based models, even though show strong behavioral generalization on appearance-free conditions, result in weak neural alignment. Together, these results indicate that world modeling motion processing approaches support robust appearance-free decoding and provide the closest overall match to the temporal dynamics of IT population representations.

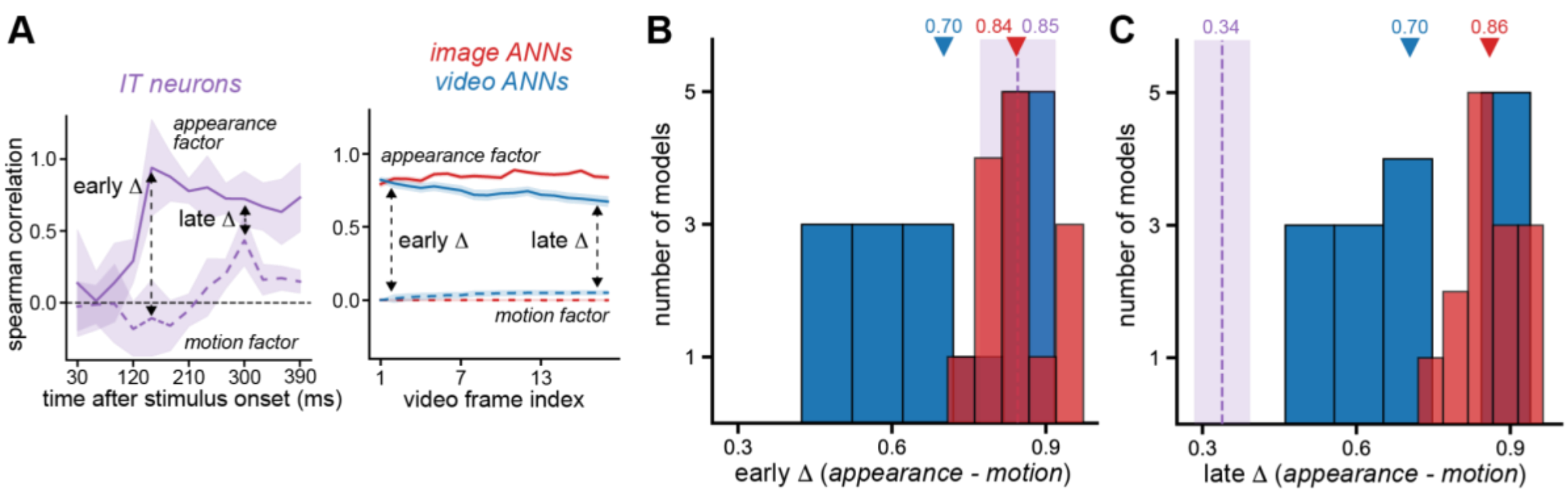

**Figure S4: Disentangling appearance and motion factors in IT neurons and image and video ANN representations. A.** Time-resolved evolution of appearance and motion factors. Left: population responses of inferior temporal (IT) neurons (N = 122 sites) show a strong early dominance of appearance-driven information, followed by a gradual increase of motion-driven signals at later time points, indicating the emergence of motion representations that are partially invariant to object appearance. At each time bin, median and standard error over neurons are reported. Right: ANN features (1000 per model), from image-based (red) and video-based (blue) architectures, exhibit largely stable contributions of appearance and motion factors over time, with appearance information consistently dominating and little evidence of a late increase in appearance-invariant motion encoding. At each video frame, median and median absolute deviation over units are reported. **B–C.** Distribution across systems of the difference between appearance and motion components (appearance Spearman − motion Spearman), comparing IT neurons (purple), image-based ANNs (red), and video-based ANNs (blue). **B.** Early response period. All systems are dominated by appearance information, with comparable appearance–motion differences across IT and ANN models (IT: 0.85, image ANNs: 0.84, video ANNs: 0.70). The purple dashed line and shaded area report median and median absolute deviation over neurons, respectively. For ANN models, the histogram of medians over units is shown. **C.** Late response period. The same descriptive statistics used in (**B**) are reported. IT neurons show a substantial reduction in the appearance–motion gap (0.34), reflecting the emergence of dynamic, appearance-invariant motion representations. In contrast, ANN models retain a stronger bias toward static

appearance information, with image-based ANNs showing the largest gap (0.86). Overall, while video-based architectures introduce temporal processing, they fail to reproduce the progressive reweighting from appearance to motion observed in the IT cortex, indicating a key discrepancy in how biological and artificial systems integrate dynamic visual information.

**A**

| Task | Model Name | IT-like Layer Name | Architecture | Training Dataset |
|---|---|---|---|---|
| Object Recognition | AlexNet | features.12 | 2d convolution | IN |
| | ConvNeXt | stages.2.blocks.4.conv_dw | hybrid | IN |
| | ConvRNN | conv10 | recurrent | IN |
| | CORnet-S | IT | recurrent | IN |
| | DenseNet-121 | features.transition3.pool | 2d convolution | IN |
| | EfficientNet-b0 | blocks.6.0 | 2d convolution | IN |
| | Hiera | blocks.7.norm2 | transformer | IN |
| | Inception-v3 | Mixed_7a | 2d convolution | IN |
| | NasNet | cell_12 | 2d convolution | IN |
| | PNASNet | cell_8 | 2d convolution | IN |
| | ResNet50 | layer4.0.relu | 2d convolution | IN |
| | ResNet50-SSL | layer4.0.relu | 2d convolution | IN |
| | ViT | blocks.6.norm2 | transformer | IN |
| | ViT-SSL | blocks.6 | transformer | IN |

**B**

| Task | Model Name | IT-like Layer Name | Architecture | Training Dataset |
|---|---|---|---|---|
| Video Recognition | C2D | blocks.4.res_blocks.5.activation | 3D convolution | K400 |
| | I3D | blocks.5.res_blocks.2.activation | 3D convolution | K400 |
| | Slow | slow_r50-blocks.4.res_blocks.2.activation | 3D convolution | K400 |
| | Slow-ssv2 | blocks.4.res_blocks.2.activation | 3D convolution | SSv2 |
| | TimesFormer | timesformer.encoder.layer.8 | transformer | K400 |
| | TimesFormer-ssv2 | timesformer.encoder.layer.11 | transformer | SSv2 |
| | VideoMAE | videomae.encoder.layer.2 | transformer | K400 |
| | VideoMamba | layers.10.mixer | state-space-model | K400 |
| | VideoMamba-ssv2 | layers.10.mixer | state-space-model | SSv2 |
| | X3D | blocks.4.res_blocks.6.activation | 3D convolution | K400 |
| Video Object Segmentation | FusionSeg | layer4 | 2D convolution, optic flow | DAVIS |
| | MatNet | layer4 | 2D convolution, optic flow | DAVIS |
| | SAM2 | memory_attention-2 | transformer | SA-V |
| World Modeling | V-JEPA2l | encoder.layer.16.mlp.fc2 | transformer | SSv2, K400, IN, YT-T-1B, H100M |
| | V-JEPA2g | encoder.layer.39.mlp.fc2 | transformer | SSv2, K400, IN, YT-T-1B, H100M |

**Figure S5: Summary of architectural specifications and training configurations for the artificial neural networks (ANNs) used in this study. A**. This table details the image-based models, all of which were trained on the ImageNet dataset for object recognition. It lists the specific internal layers designated as "IT-like" for comparison with the primate ventral stream across various architectures, including 2D convolutional, recurrent, and transformer networks. **B**. This table summarizes the video-based models, which encompass a wider range of tasks such as action recognition and video object segmentation. These models utilize diverse temporal architectures, including 3D convolutions, transformers, and state-space models, and were trained on specialized video datasets such as Kinetics-400, Something-Something-v2, DAVIS, and SA-V. For each model, the specific layer utilized for representational alignment and predictivity analysis is identified.